\documentclass[10pt,twocolumn,letterpaper]{article}

\usepackage{iccv}
\usepackage{times}
\usepackage{epsfig}
\usepackage{graphicx}
\usepackage{amsmath}
\usepackage{amssymb}
\usepackage{comment}
\usepackage{setspace} 
\usepackage{dblfloatfix}
\usepackage{booktabs}
\usepackage{multirow}
\usepackage{tablefootnote}
\usepackage[font=small]{caption}
\usepackage[dvipsnames]{xcolor}

\usepackage[subrefformat=parens]{subcaption}
\usepackage[pagebackref=true,breaklinks=true,letterpaper=true,colorlinks,bookmarks=false]{hyperref}

\usepackage[symbol]{footmisc}
\renewcommand{\thefootnote}{\fnsymbol{footnote}}

\iccvfinalcopy

\begin{document}

\title{Spatially Controllable Image Synthesis with Internal Representation Collaging\vspace{-3mm}}

\author{Ryohei Suzuki$^{1,2}$\footnotemark[2]\hspace{4mm}Masanori Koyama$^2$\hspace{4mm}Takeru Miyato$^2$\hspace{4mm}Taizan Yonetsuji$^2$\hspace{4mm}Huachun
Zhu$^2$\\\vspace{-3mm}
\\
$^1$The University of Tokyo\hspace{4mm}$^2$Preferred Networks, Inc.\\}
\twocolumn[{%
\renewcommand\twocolumn[1][]{#1}%
\maketitle
\begin{center}
    \centering
    \vspace{-6mm}
    \setlength{\abovecaptionskip}{2mm}
    \includegraphics[width=0.98\textwidth]{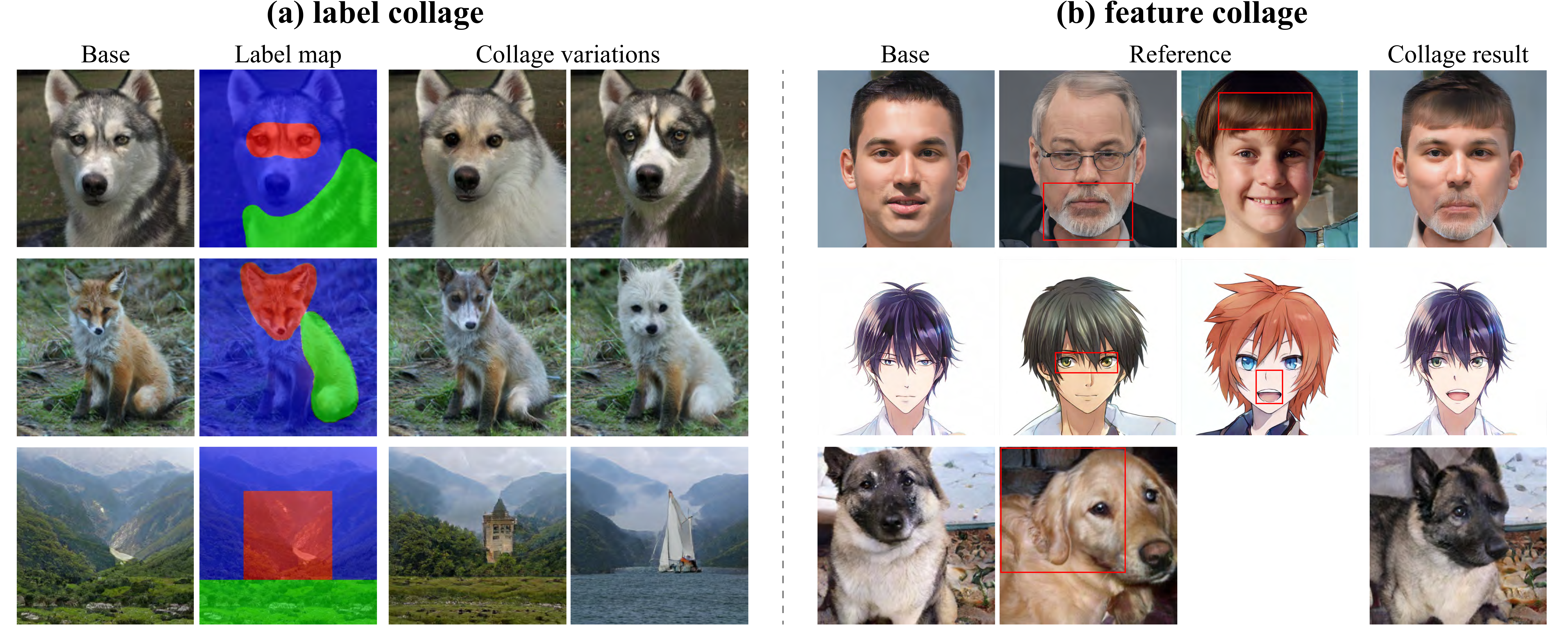}
    \captionof{figure}{Examples of \textit{label collaging} by our sCBN (a) and \textit{feature collaging} by our feature blending (b). With our sCBN, the user can change the label of the user-specified parts of an image to the user-specified target labels by specifying an appropriate label map. In each row of (a), the label information of the base image over the red colored region and the green colored region are altered by sCBN. The image at the right bottom corner is the result of changing the label of the red region to \textit{yawl} and changing the label of the green region to \textit{lakeside}. The images in the right-most column of the panel~(b) are the results of our feature blending method. In each row, the red-framed regions in the reference images are blended into the base image. In the second row, eye features of the left reference and mouth features of the right reference are blended into the base male image. Our methods can be applied to a wide variety of images. 
    }
    \label{fig:teaser}
\end{center}%
}]

\begin{abstract}
We present a novel CNN-based image editing strategy that allows the user to
change the semantic information of an image over an arbitrary region by manipulating the feature-space representation of the image in a trained GAN model. 
We will present two variants of our strategy:  (1) spatial conditional batch normalization (sCBN), a type of conditional batch normalization with user-specifiable spatial weight maps, and (2) feature-blending, a method of directly modifying the intermediate features.
Our methods can be used to edit both artificial image and real image, and they both can be used together with any GAN with conditional normalization layers. 
We will demonstrate the power of our method through experiments on various types of GANs trained on different datasets.
Code will be available at \url{https://github.com/pfnet-research/neural-collage}.
\footnotetext[2]{This work was done when the author was at Preferred Networks, Inc.}
\end{abstract}
\newpage

\renewcommand*{\thefootnote}{\arabic{footnote}}


\section{Introduction}

Deep generative models like generative adversarial networks (GANs)~\cite{goodfellow2014generative} and variational autoencoders (VAEs) are powerful techniques for the unsupervised learning of latent semantic information that underlies the data.  
GANs has been particularly successful on the tasks on images: applications include image colorization~\cite{iizuka2016let,isola2017image}, inpainting~\cite{pathak2016context,iizuka2017globally,yu2018generative}, domain translation~\cite{isola2017image,yi2017dualgan,zhu2017toward,sangkloy2017scribbler,wang2018high}, style transfer~\cite{huang2017arbitrary,zhang2017multi}, object transfiguration~\cite{zhu2017unpaired,liu2017unsupervised,huang2018multimodal,liang2018generative}, just to name a few.
Generation of photo-realistic images with large diversity has also been made possible by the invention of techniques to stabilize the training process of GANs over massive datasets~\cite{miyato2018spectral,gulrajani2017improved} as well \cite{zhang2018self,karras2017progressive, miyato2018cgans, brock2018large, karras2018style}.
 
But the challenge remains to regulate the GANs' output at the user's will. 
In one of the earliest attempts on this problem, conditional GAN~\cite{mirza2014conditional} established the strategy of concatenating the latent input vector with a vector that represents conditional information.
One can change the semantic information of the image by manipulating the conditional vector. 
InfoGAN~\cite{chen2016infogan} further extended the idea by modeling the disentangled latent features with a set of independent conditional vectors and optimizing the model with information theoretic penalty term. 
More recently, StyleGAN~\cite{karras2018style} succeeded in automatically isolating the high-level attributes and in realizing a scale-specific control of the image synthesis.

Meanwhile, in creative tasks, for example, it often becomes necessary to transform just small regions of interest in an image.
Many methods in practice today use annotation datasets of semantic segmentation.
These methods include~\cite{isola2017image,wang2018high,park2019spade}, which can be used to construct a photo-realistic image from a doodle (label map). 
To author's best knowledge, however, not much has been done for the unsupervised transformation of images with spatial freedom. 
Recently, GAN dissection~\cite{bau2018gan} explored the semantic relation between the output and the intermediate features and succeeded in using the inferred relation for photo-realistic transformation. 
In this paper, we present a strategy of image transformation that is strongly inspired by the findings in~\cite{bau2018gan}.  
Our strategy is to manipulate the intermediate features of the target image in a trained generator network. 
We present a pair of novel methods based on this strategy---Spatial conditional batch normalization and Feature blending---that apply affine transformations on the intermediate features of the target image in a trained generator model.
Our methods allow the user to \textit{edit} the semantic information of the image in a \textit{copy and paste} fashion.

Our Spatial conditional batch normalization is a spatial extension of conditional normalization~\cite{de2017modulating,perez2017film,huang2017arbitrary}, and it allows the user to blend the semantic information of multiple labels based on the user-specified spatial map of mixing-coefficients (\textit{label collaging}).
With sCBN, we can not only generate an image from a label map but also make local semantic changes to the image like \textit{changing the eyes of a husky to eyes of a Pomeranian}
(Fig.~\ref{fig:teaser}a).
On the other hand, our Feature blending is a method that directly mixes multiple images in the intermediate feature space, and it enables local blending of more intricate features (\textit{feature collaging}).
With this technique, we can make modifications to the image like \textit{changing the posture of an animal} without providing the model with the explicit definition of the \textit{posture}
(Fig.~\ref{fig:teaser}b).

\begin{figure}[t!]
   \centering
   \includegraphics[width=2.1in]{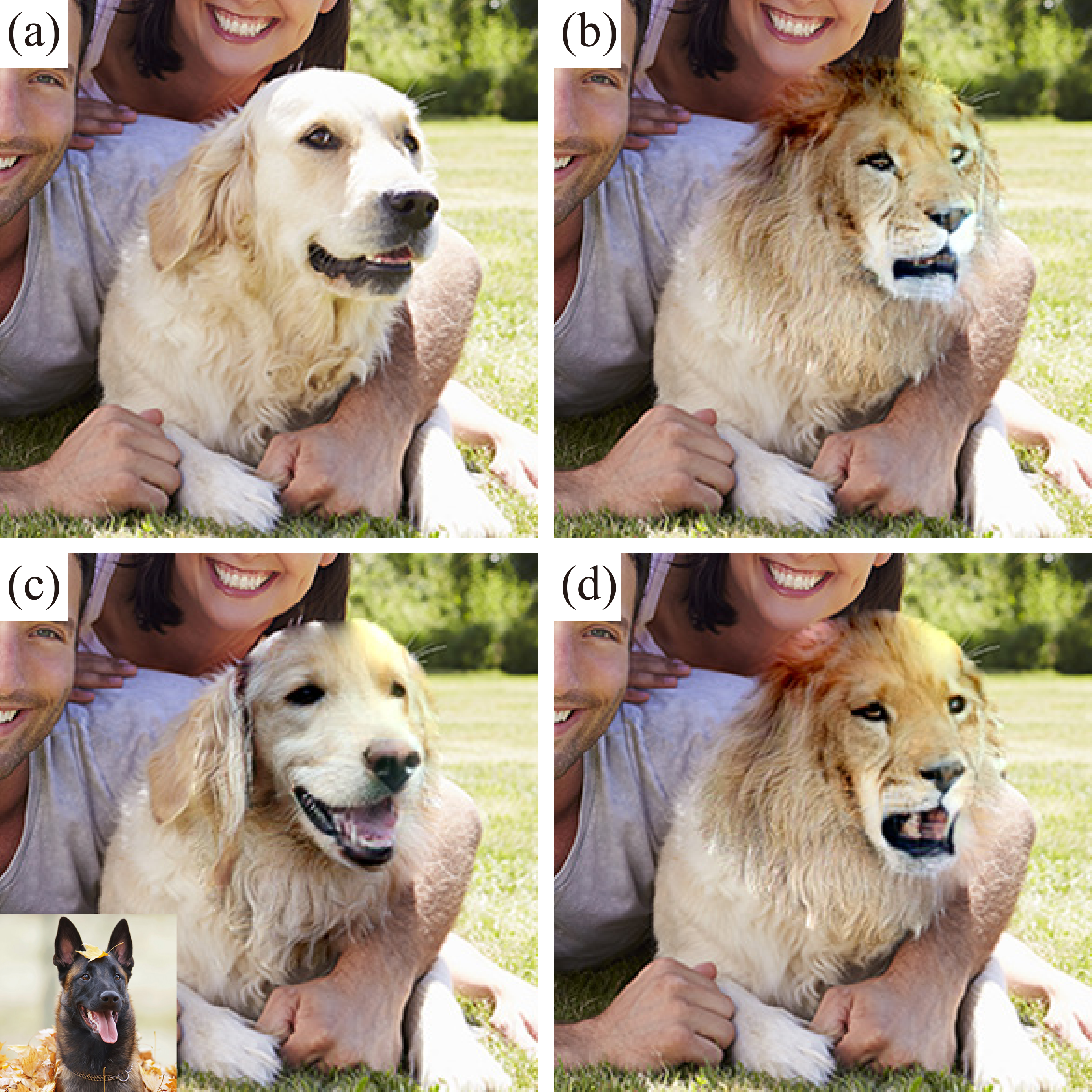}
   \caption{Application of our techniques to a real image. (a) input \textit{real} image, (b) label collaging by sCBN (Retriever to Lion), (c) feature blending (closed mouth to open mouth), and (d) a combination of the two (a Retriever with closed mouth to a Lion with open mouth). The small image located at the corner of (c) is the source of the blended feature (open mouth).} 
   \label{fig:intro_editing}
 \end{figure}

One significant strength common to both our methods is that they only require a trained GAN that is equipped with AdaIN/CBN structure; there is no need to train an additional model.
Our methods can be applied to practically any types of images for which there is a well-trained GAN. 
Both methods can be used together as well to make an even wider variety of semantic manipulation on images. 
Also, by combining our methods with manifold projection~\cite{zhu2016generative}, we can manipulate the local semantic information of a \textit{real} image (Fig. ~\ref{fig:intro_editing}). 
Our experiments with the most advanced species of GANs~\cite{miyato2018cgans,brock2018large,karras2018style} shows that our strategy of ``standing on the shoulder of giants" is a sound strategy for the task of unsupervised local semantic transformation.  

\section{Related Work}

In this section, we will present the previous literature that are closely related to our study. 

\paragraph{Generative adversarial networks (GANs).}
GAN is a deep generative framework consisting of a generator function $G$ and a discriminator function $D$ that play a min-max game in which $G$ tries to transform the prior distribution \eg, $\mathcal{N}(0, I)$ into a good imposter distribution that mimics the target distribution and $D$ tries to distinguish the artificially generated data from the true samples~\cite{goodfellow2014generative}.
Thanks to the development of regularization techniques like gradient penalty~\cite{gulrajani2017improved} and spectral normalization \cite{miyato2018spectral}, deep convolutional GANs \cite{radford2015unsupervised} are becoming a de facto standard for many image generation tasks.
GANs also excel in representation learning, and it is known that one can continuously transform the image in a somewhat semantically meaningful way by interpolating between a pair of 
points in the latent space~\cite{radford2015unsupervised,brock2016neural,odena2016conditional,brock2018large}.

\paragraph{Conditional normalization layers}

Classic conditional GAN used to incorporate class-specific information into the output by concatenating a class embedding vector to the latent variable. 
Becoming more popular in recent years is the strategy of inserting a conditional normalization layer into a network~\cite{miyato2018cgans,brock2018large,karras2018style,park2019spade}. 
Conditional batch normalization (CBN)~\cite{de2017modulating,perez2017film} is a mechanism that can learn conditional information in forms of class-specific scaling parameter and shifting parameter.
Also belonging to the same family is an adaptive instance normalization (AdaIN) that learns output-conditional scaling parameter and shifting parameter.
By manipulating the parameters of the conditional normalization layers, one can manipulate the semantic information of the images in impressively natural fashion. 

SPADE~\cite{park2019spade} is also a method based on conditional normalization that was developed almost simultaneously with our method, and it can learn a function that maps an arbitrary segmentation map to an appropriate parameter map of the normalization layer that can convert the segmentation map to a photo-realistic image.
Naturally, the training of the SPADE-model requires a dataset contains annotated segmentation map. 
However, it can be a nontrivial task to obtain a generator model that is well trained on a dataset of annotated segmentation maps for the specific image-type of interest,  let alone the dataset of annotated segmentation map itself. 
Our method makes a remedy by taking the approach of modifying the conditional normalization layers of a \textit{trained} GAN.
Our \textit{spatial conditional batch normalization} (sCBN) takes a simple strategy of applying position-dependent affine-transformations to the normalization parameters of a trained network, and it can spatially modify the semantic information of the image without the need of training a new network. 
Unlike the manipulation done in style transfer~\cite{huang2017arbitrary}, we can also edit the conditional information at multiple levels in the network and control the effect of the modification.
As we will investigate further in the later section, modification to a layer that is closer to the input tends to transform more global features.

\paragraph{Direct manipulation of intermediate representations}

Sometimes, the feature to be transplanted/transformed does not correspond to specific labels/classes. 
That a specific semantic feature of an image often corresponds to a specific set of neurons in CNN is a fact that has been known from long ago~\cite{le2011building}.
Numerous approaches have taken advantage of this property of CNN to transfer the styles/attributes of one image to another~\cite{gatys2015neural,upchurch2017deep,Liao:2017:VAT:3072959.3073683}.
Very recently, GAN dissection~\cite{bau2018gan} took a very systematic approach that utilizes the correlation between the intermediate features and semantic segmentation map.
By identifying the set of intermediate features that correspond to the semantic feature of interest, GAN dissection succeeded in making modifications to an image like ``increasing the number of trees in the park."
One particular advantage of the strategy taken by~\cite{bau2018gan} is that the user does not need to explicitly define the feature that he/she wants to modify in the image when training the generator. 
Our method is inspired by the findings of~\cite{bau2018gan} and utilizes the conditional information that has already been learned by a trained generator. 
By blending the intermediate features of multiple images with spatially varying mixture-coefficients, the user can make a wide variety of artificial images without compromising the photo-realism.

Finally, by combining our methods with manifold projection~\cite{zhu2016generative}, we can adopt the strategy similar to the ones taken by~\cite{zhu2016generative,brock2016neural,kaneko2017generative} and make spatial semantic manipulation of \textit{real} images as well.


\section{Two Methods of Collaging the Internal Representations}
The central idea common to both our sCBN and Feature blending is to modify the intermediate features of the target image in a trained generator using a user-specifiable masking function. 
In this section, we will describe the two methods in more formality. 

\subsection{Spatial Conditional Batch Normalization}
\begin{figure}[t!]
\centering

\includegraphics[height=1.85in]{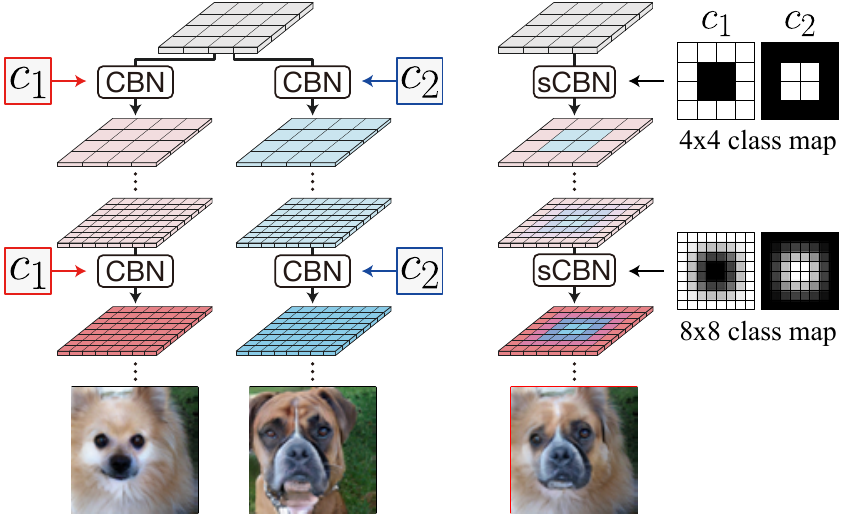}

\caption{Schematic comparison of CBN and sCBN layers; CBN layers gradually introduce the class-specific features into the generated image with spatially uniform strength (left). sCBN layers do practically the same thing, except that they apply the class-specific features with user-specified mixing intensities that vary spatially across the image.}
\label{fig:spatialcbn}
\end{figure}

As can be inferred from our naming, spatial conditional batch normalization~(sCBN) is closely related to conditional batch normalization~(CBN) \cite{dumoulin2017learned,de2017modulating}, a variant of Batch Normalization that encodes the class-specific semantic information into the parameters of BN.
For locally changing the class label of an image, we will apply spatially varying transformations to the parameters of conditional batch normalization~(sCBN) (Fig.~\ref{fig:spatialcbn}). 
Given a batch of images sampled from a single class, the conditional batch normalization \cite{dumoulin2017learned,de2017modulating} 
normalizes the set of intermediate features produced from the batch by a pair of class-specific scale and bias parameters.
Let $F_{k, h, w}$ represent the feature of $\ell$-th layer at the channel $k$, the  height location $h$, and the width location $w$.
Given a batch $\{F_{i, k, h, w}\}$ of $F_{k, h, w}$s generated from a class $c$, the CBN at layer $\ell$ normalizes $F_{i, k, h, w}$ by:
\begin{equation}
    \hat{F}_{i,k,h,w} \leftarrow
    \gamma_{k}(c) \frac{F_{i,k,h,w} - \mathrm{E}\left[F_{\cdot,k,\cdot,\cdot}\right]}{\sqrt{\mathrm{Var}\left[F_{\cdot,k,\cdot,\cdot}\right] + \epsilon}} + \beta_{k}(c) \label{eq:CBN}
\end{equation}
where $\gamma_{k}(c), \beta_{k}(c)$ are respectively the trainable scale and bias parameters that are specific to the class $c$.
The sCBN modifies \eqref{eq:CBN} by replacing $\gamma_k(c)$ with $\tilde{\gamma}_{k,h,w}$ given by
\begin{equation}
    \tilde{\gamma}_{k,h,w} := \sum_{c=1}^{N_{\mathrm{class}}}
    W_{h,w}(c) \cdot \gamma_k(c)
\end{equation}
where $W_{h,w}(c)$ is a user-selected non-negative tensor map (class-map) of mixture coefficients that integrates to $1$; that is, $\sum_{c} W_{h,w}(c) = 1$ for each position $(h,w)$.
We apply an analogous modification on $\beta_k(c)$ to produce $\tilde{\beta}_{k,h,w}$.
If the user wants to modify an artificial image generated by the generator function $G$, the user may replace the CBN of $G$ at (a) user-chosen layer(s) with sCBN with (a) user-chosen weight map(s).
The region in the feature space that will alter the user-specified region of interest can be inferred with relative ease by observing the downsampling relation in $G$. 
The user can also control the intensity of the feature of the class $c$ at an arbitrary location $(h,c)$ by choosing the value of $W_{h,w}(c)$ (larger the stronger). 
By choosing $W$ to have strong intensities for different classes in different regions, the user can transform multiple disjoint parts of the images into different classes~(see figure~\ref{fig:teaser}a).
As we will show in section \ref{exp:layers}, the choice of the layer(s) in $G$ to apply sCBN have interesting effects on the intensity of the transformation. The figure \ref{fig:spatialcbn} shows the schematic overview of the mechanism of sCBN. 
By using manifold projection, sCBN can be applied to real images as well.
We will elaborate more on the application of our method to real images in section \ref{sec:app}.

\subsection{Spatial Feature Blending}
\begin{figure}[t!]
\centering

\includegraphics[height=2.05in]{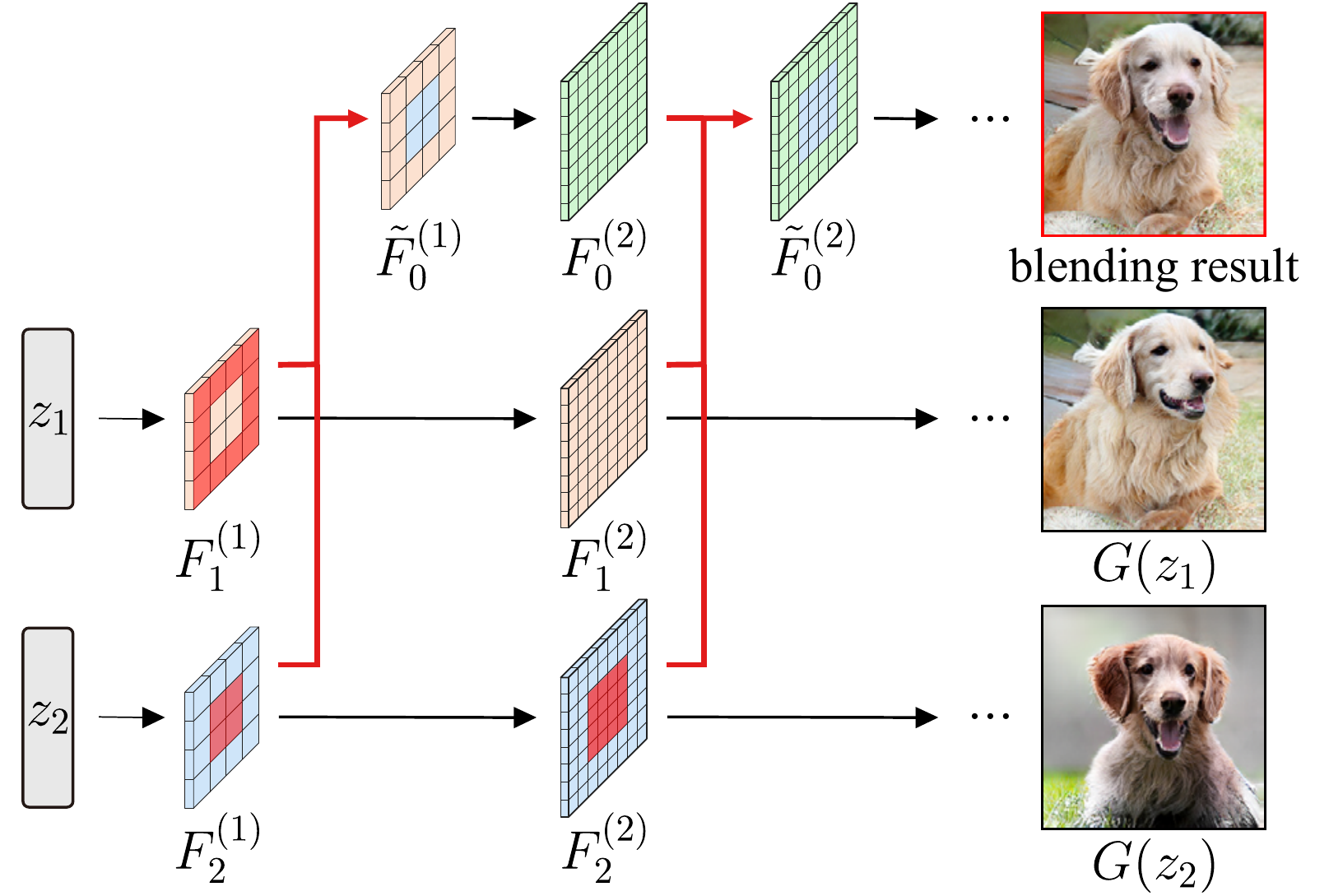}

\caption{An overview of spatial feature blending. Reference images generated from different latent variables are iteratively mixed into the target image in the feature space of a trained generator.}
\label{fig:featblend}
\end{figure}
Our spatial feature blending is a method that can extract the features of a particular part of one image and blend it into another. 
Suppose that images $x_i$ are generated from latent variables $z_i$ by a trained generator $G$,  
and that ${F}_i^{(\ell)}$ are the
feature map of the image $x_i$ that can be obtained by applying $\ell$ layers of $G$ to $z_i$. 
We can then blend the features of $x_i; i>0$ into $x_0$ by  
recursively replacing ${F}^{(\ell)}_0$ with 
\begin{equation}
    \tilde{F}^{(\ell)}_0 := \sum_{i=0}^{N_{\mathrm{input}}} M^{(\ell)}_i \odot U^{(\ell)}_i(F^{(\ell)}_i),
\end{equation}
where $\odot$ is the Hadamard product, $M^{(\ell)}_i$ is the user-specified non-negative tensor (feature blending weights) with the same dimension as $F^{(\ell)}_i$ such that $\sum_i M^{(\ell)}_{i}$ is the tensor whose entries are all $1$, and $U^{(\ell)}_i$ is an optional shift operator that uniformly translate the feature map $F^{(\ell)}_i$ to a specified direction, which can be used to move a specific local feature to an arbitrary position.

As a map that is akin to the class map $W^{(\ell)}_i$ in sCBN, the user may choose $M^{(\ell)}_i$ in a similar way as in the previous section to spatially control the effect of the blending. 
Spatial feature blending can also be applied to real images by using the method of manifold projection. 
The figure \ref{fig:featblend} is an overview of the feature-blending process in which the goal is to transplant a feature (front facing open mouth) of an image $G(z_2)$ to the target image $G(z_1)$(a dog with a closed mouth).
All the user has to do in this scenario is to provide a mixing map $M$ that has high intensity on the region that corresponds to the region of the mouth.
As we will show in the experiment section, our method is quite robust to the alignment, and the region of mouth in $G(z_2)$ and $G(z_1)$ needs to be only roughly aligned.

\section{\label{sec:app}Application to Real Images}
\begin{figure}[t]
\centering

\includegraphics[height=1.4in]{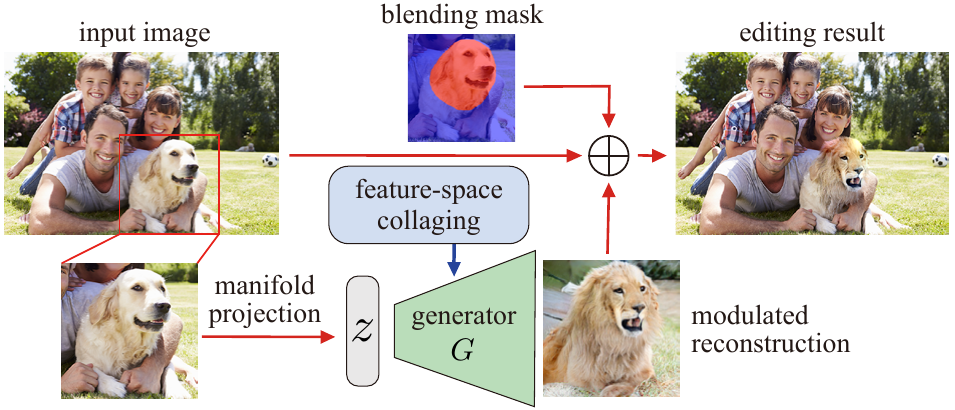}

\caption{An overview of the algorithm of applying the feature-space collaging to the real image. The user is asked to specify the blending map, choice of the feature-space collaging method, and the mask to be used for the Poisson blending in the post-processing step.}
\label{fig:algo-overview}
\end{figure}

To edit the semantic information of a real image, we can use the method of manifold projection~\cite{zhu2016generative} that looks for the latent variable $z$ such that 
$G(z) \cong x$.
After obtaining the \textit{inverse} of $x$, we can apply the same set of procedures we described above to either modify the label information of parts of $x$  or blend the features of other images into $x$ or do both.  
The figure \ref{fig:algo-overview} is the flow of real-image editing.

\begin{figure}[t!]
\centering

\includegraphics[height=2.25in]{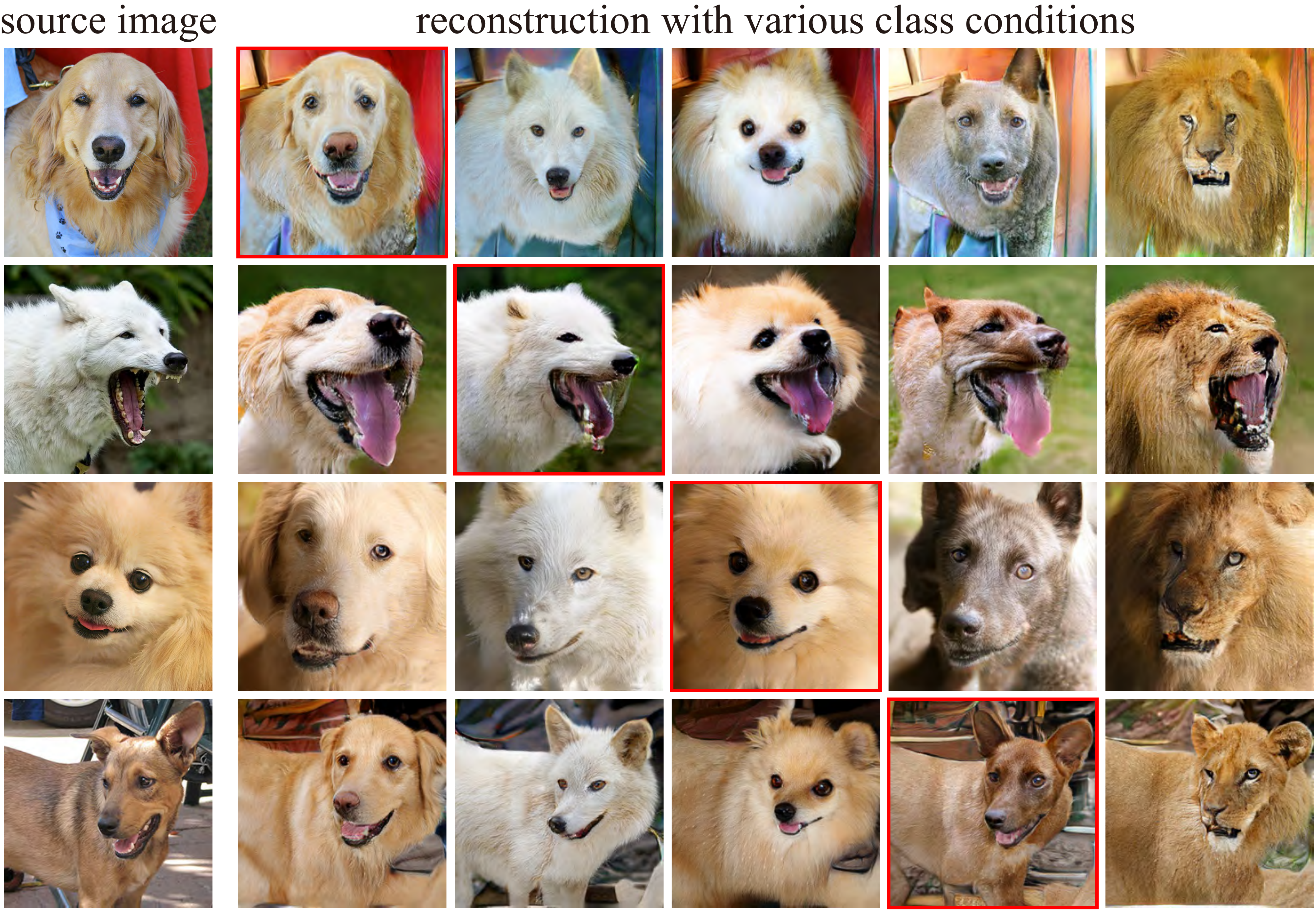}

\caption{Examples of image reconstruction by manifold projection with various class labels. The images with red frames are the reconstruction with the original class label.}
\label{fig:nonspatial}
\end{figure}

We will describe this process in more formality. 
Let $(G,D)$ be the pair of a trained generator and a discriminator.
Given an image $x$ of interest (often a clip from a larger image), the first goal of the manifold projection step is to train the encoder $E$ such that $\mathcal{L}(G(E(x)), x)$ is small for some 
dissimilarity measure $\mathcal{L}$.
The choice of $\mathcal{L}$ in our method is the cosine distance in the final feature space of the discriminator $D$. That is, if $\mathbf{\Psi}(x)$ is the normalized feature of the image $x$ in the final layer of $D$,
\begin{equation}
    \mathcal{L}(x_1, x_2) := 1 - \mathbf{\Psi}(x_1) \cdot \mathbf{\Psi}(x_2). \label{eq:cosine}
\end{equation}
After training the encoder, one can produce the reconstruction of $x$ by applying $G$ to $z = E(x)$. In the reconstructed image, however, semantically independent objects can be dis-aligned.  We, therefore, calibrate $z$ by additionally backpropagating the loss $\mathcal{L}$. After some rounds of calibration, we can feed the obtained $z$ to the modified $G$ to create a transformed reconstruction. 
Please see the supplementary material for the details of the manifold projection algorithm.
The figure \ref{fig:nonspatial} shows examples of reconstruction with various class conditions.

As a final touch to the transformed image, we apply a post-processing step of Poisson blending \cite{perez2003poisson}.
Because most trained GAN models do not have the ability to disentangle objects from the background, naively pasting the generated clip to the target image can produce artifacts in the region surrounding the object of interest.
We can clean up these artifacts with Poisson blending to the region of interest.

\begin{figure}[t!]
\centering

\includegraphics[height=1.6in]{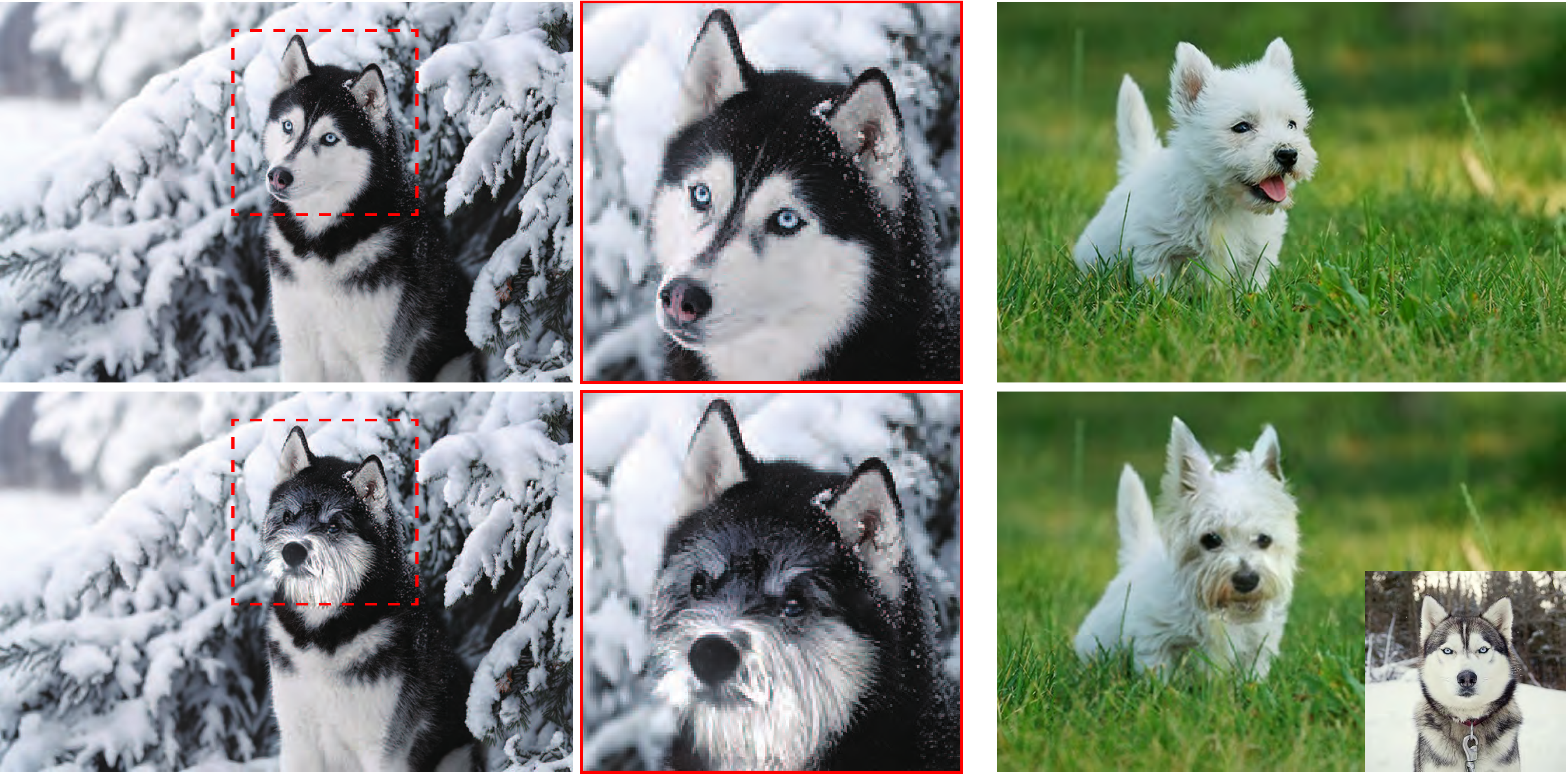}

\caption{Left panel: an example of label collaging by sCBN applied to a real image. The clip enclosed in a dashed red frame is the region that was transformed.   Right panel: an example of feature blending applied to a real image.  We changed the head position of a white terrier by blending the feature of the front-facing husky.}
\label{fig:edit-results}
\end{figure}

Fig.~\ref{fig:edit-results} are examples of the application of our methods to real images.
The image in the left panel is an application of sCBN to a real image, and the image in the right panel is an application of feature blending to a real image.

\section{Experiments}

In this section, we present the results of our experiments together with the experimental settings. For further details, please see the supplementary material.

\subsection{Experimental Settings}
We applied our methods to ResNet-based generators that were trained as parts of conditional GANs:  SNGAN~\cite{miyato2018spectral, miyato2018cgans},  BigGAN~\cite{brock2018large}, and StyleGAN~\cite{karras2018style}.
These GANs are all equipped with conditional normalization layers. 
In the experiments, we treated the AdaIN layer in the StyleGAN in the same manner as the conditional batch normalization layer. 
Both SNGAN and BigGAN used in our study first map the latent vector into feature maps of dimension $4 \times 4$, and doubles the resolution at every ResBlock to produce an RGB image of the user-specified dimensions in the final layer.
We produced both $128 \times 128$ (for Flowers dataset) and $256 \times 256$ (for Dogs+Cats dataset) images with SNGAN, and produced   $256 \times 256$ images with BigGAN.
The StyleGAN used in this study produces the image of 1024 $\times$ 1024 resolution by recursively applying convolution layers and AdaIN layers to the latent vector. 

For the input of $\mathbf{\Psi}$ in the equation \eqref{eq:cosine} for the transformation of $128 \times 128$~($256 \times 256$) dimensional real image, we used $512~(768)$ dimensional feature vectors of the discriminator $D$ prior to the final global pooling, which are known to capture the semantic features of the image well ~\cite{dosovitskiy2016generating,johnson2016perceptual}.

\subsection{Results} 
\begin{table}[]
    \centering
    \begin{tabular}{ll}
    \textbf{Dataset} & \textbf{Model}  \\
    \toprule
    $^*$Dogs+Cats (143 classes)\footnotemark[1] & \multirow{ 2}{*}{SNGAN~\cite{miyato2018spectral, miyato2018cgans}}\\
    Flowers (102 classes) \cite{Nilsback08} & \\
    \midrule
    $^*$ImageNet (1000 classes)~\cite{ILSVRC15} & BigGAN~\cite{brock2018large}\\
    \midrule
    $^*$FFHQ\footnotemark[2] & \multirow{2}{*}{StyleGAN~\cite{karras2018style}} \\
    Danbooru (anime face)\footnotemark[3] &  \\
    \bottomrule
    \end{tabular}
    \caption{List of the model-dataset pairs to which we applied our algorithm. We directly used the publicized implementations for the starred models.}
    \label{tab:models_and_datasets}
\end{table}

We demonstrate the utility of our algorithm with several DCGANs trained on different image datasets (see figure~\ref{tab:models_and_datasets}).
In order to verify the representation power of our manifold projection method and the DCGANs used in our study, we conducted a set of ablation experiments for \textit{non-spatial} transformation as well. 
We confirmed that the generators used in our study are powerful enough to capture the class-invariant intermediate features (see figure~\ref{fig:nonspatial} and the supplementary material).
\footnotetext[1]{subset of ImageNet~\cite{ILSVRC15}}
\footnotetext[2]{\url{https://github.com/NVlabs/ffhq-dataset}}
\footnotetext[3]{\url{https://www.gwern.net/Danbooru2018}}

\begin{figure}[t]
\centering
\includegraphics[width=3.25in]{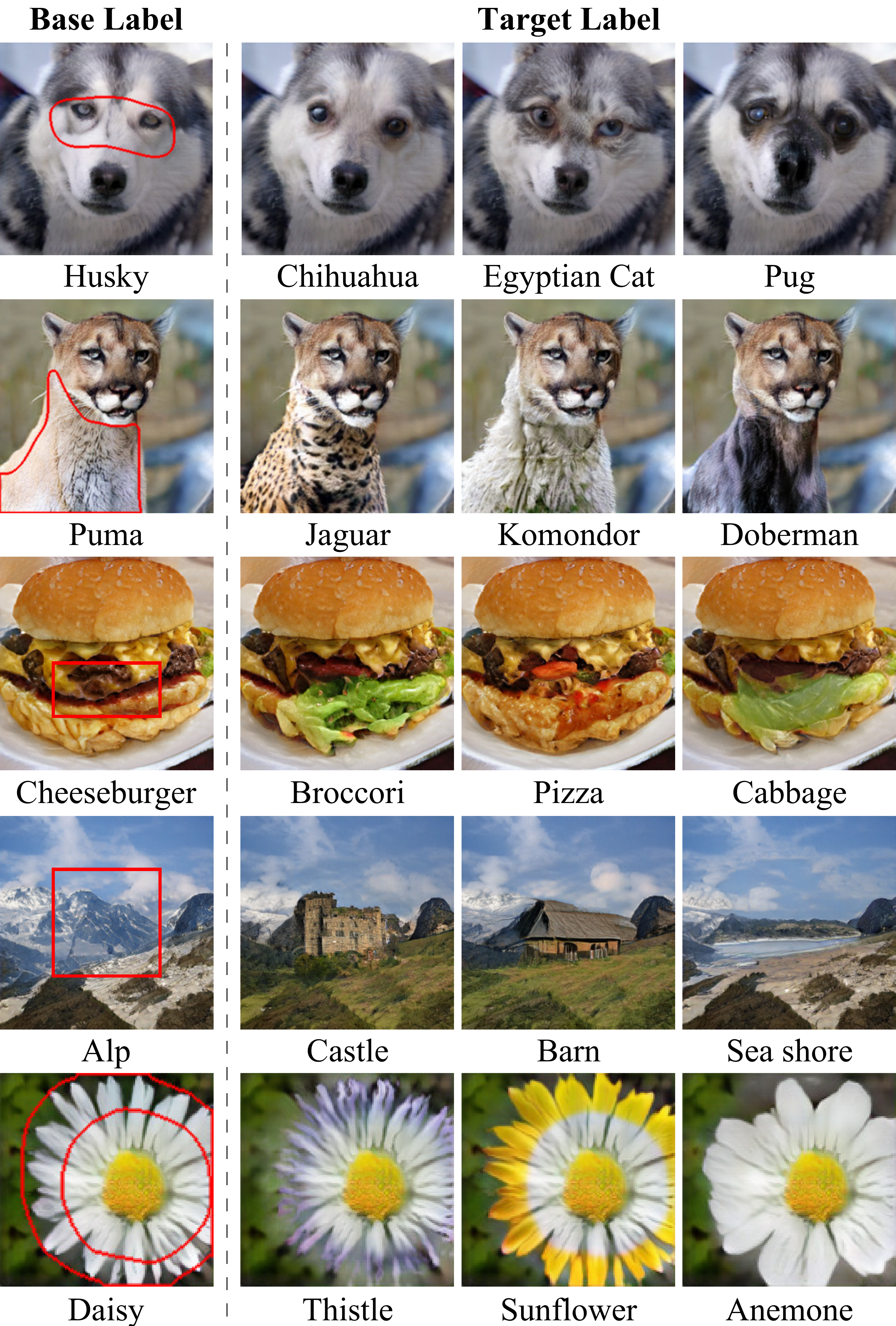}
\caption{Examples of label collaging by sCBN. The region enclosed by the red line was translated to the target label. Top two rows and the bottom rows are the results of applying our method to SNGAN. Other rows are the results of our method on BigGAN. See table~\ref{tab:models_and_datasets} for the dataset used for training the generator.}
\label{fig:exp-label-collage}
\end{figure}

\begin{figure}[t]
\centering
\includegraphics[width=3.25in]{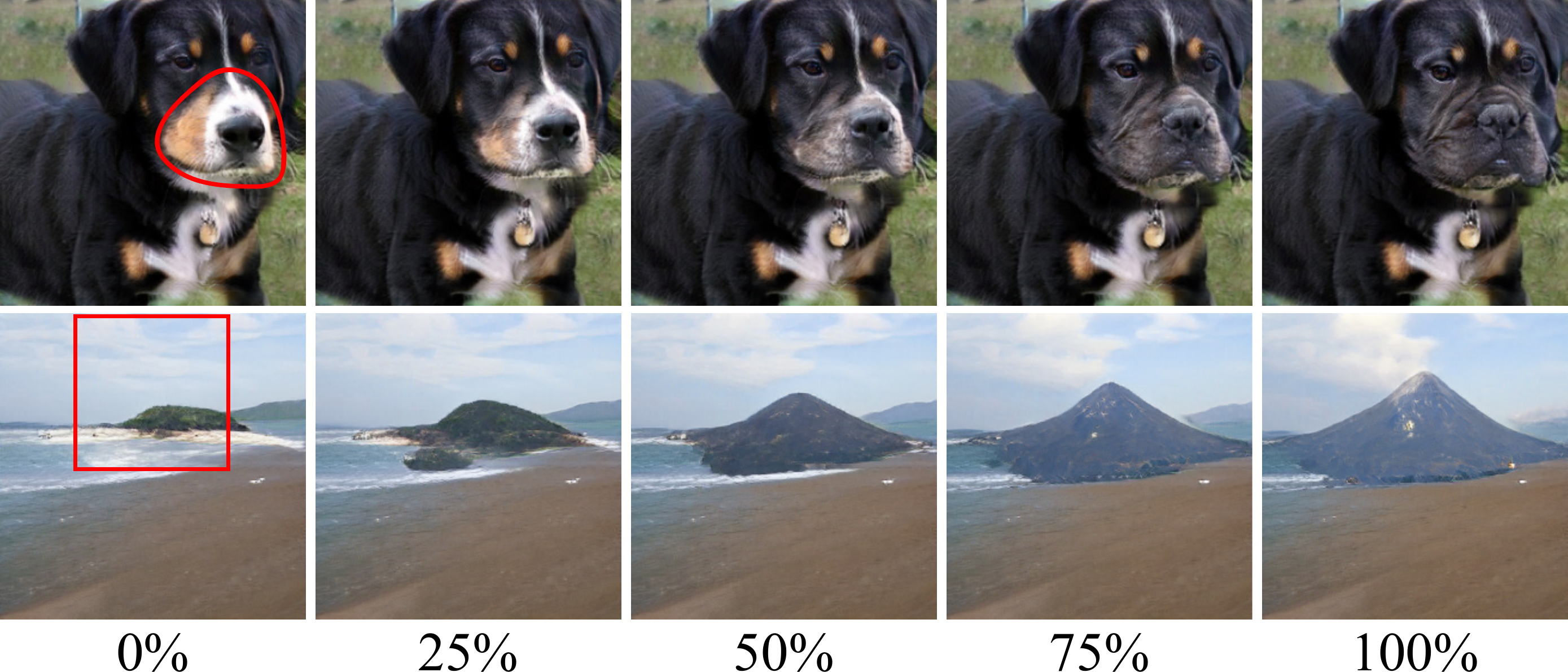}
\caption{Examples of label collaging with different intensity.  The region enclosed by the red line was translated to a target label with the shown intensities.}
\label{fig:exp-label-collage-interp}
\end{figure}

The images in figure \ref{fig:exp-label-collage} are examples of our label collaging with sCBN.
We see that our transformation is modulating both global information~(e.g.~face/shape) and local information~(e.g.~color/texture) without breaking the semantic consistency.
Also, note that areas outside the specified region are also affected when global features like ``face of a dog" are altered. 
Interestingly, our algorithm is automatically choosing the region to be influenced by the modification of global feature so that there will not be an unnatural discontinuity in the transformed image. 
Because the user-specified region of interest is defined in the pixel space, this observation suggests that the algorithm is modulating the global features in the layers close to the input.
Because the user is allowed to assign continuous values to the label map, the intensity of collaging can be continuously controlled as well (see figure~\ref{fig:exp-label-collage-interp}). 

\begin{figure}[t]
\centering
\includegraphics[width=3.25in]{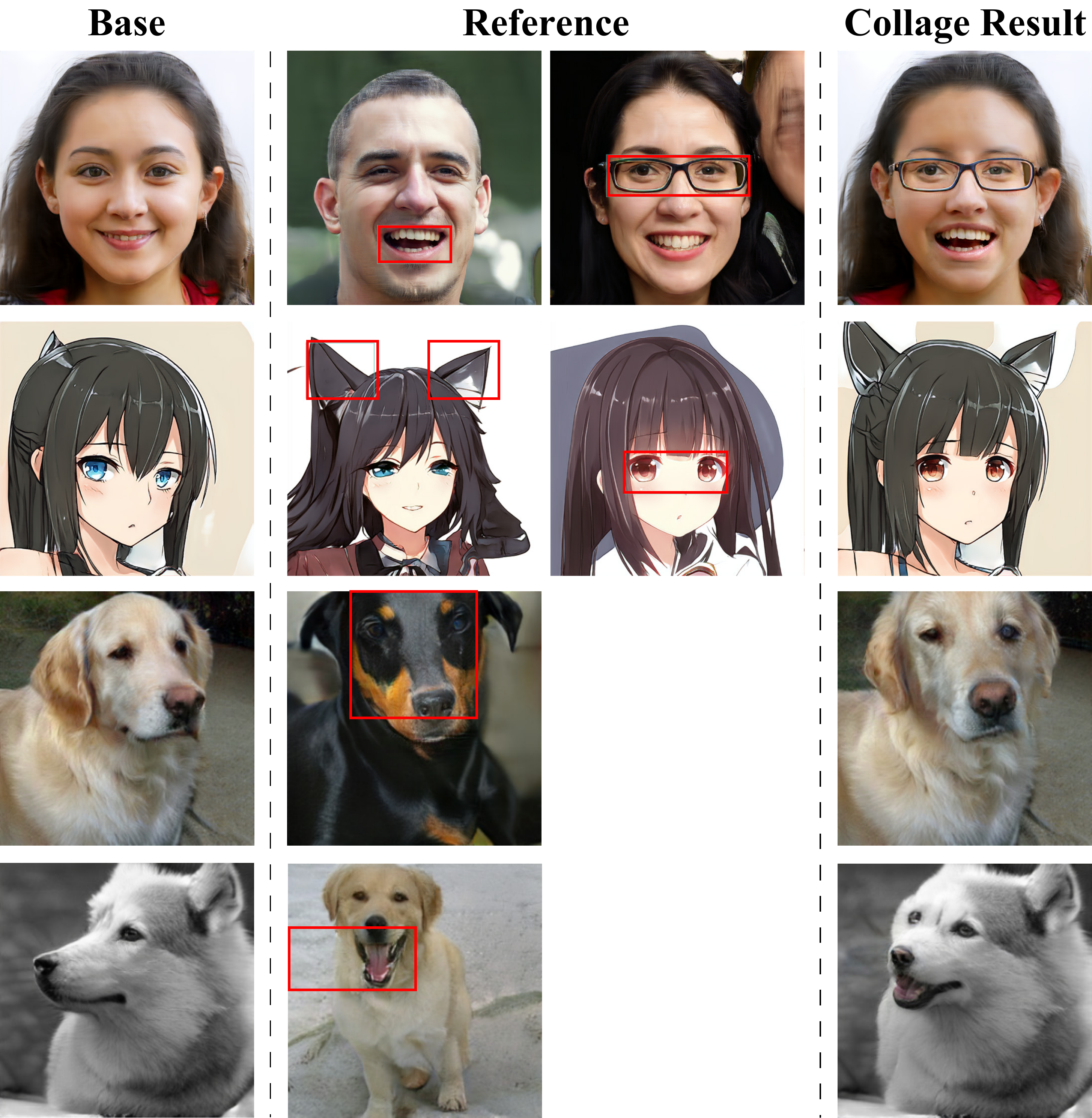}
\caption{Examples of feature collaging. Features inside the regions in the red frame are blended into the base image. Top two rows are the result of applying our method to StyleGAN. Other rows are the results of our method on SNGAN. Note that our method performs well on various types of dataset.}
\label{fig:exp-feature-collage}
\end{figure}

\begin{figure}[t]
\setlength{\abovecaptionskip}{0.10in}
\centering

\includegraphics[width=3.25in]{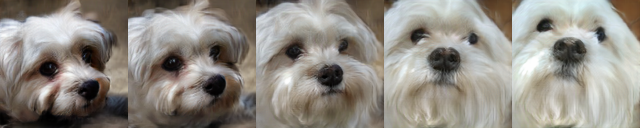}\\
\includegraphics[width=3.25in]{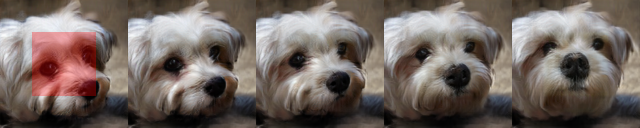}\\
\vspace{0.01in}
\includegraphics[width=3.25in]{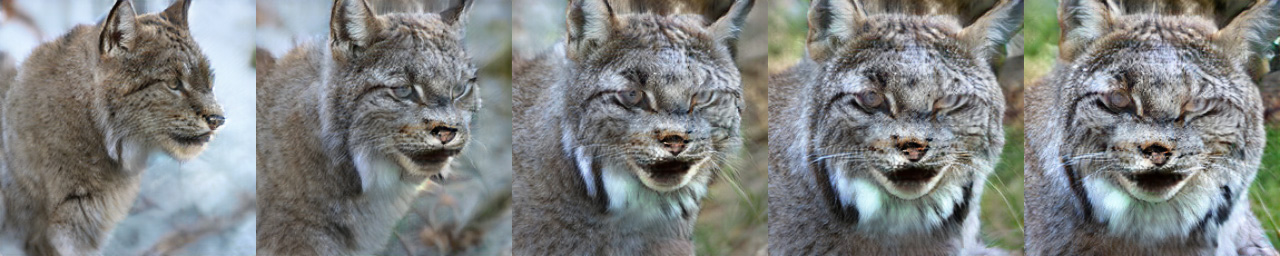}\\
\includegraphics[width=3.25in]{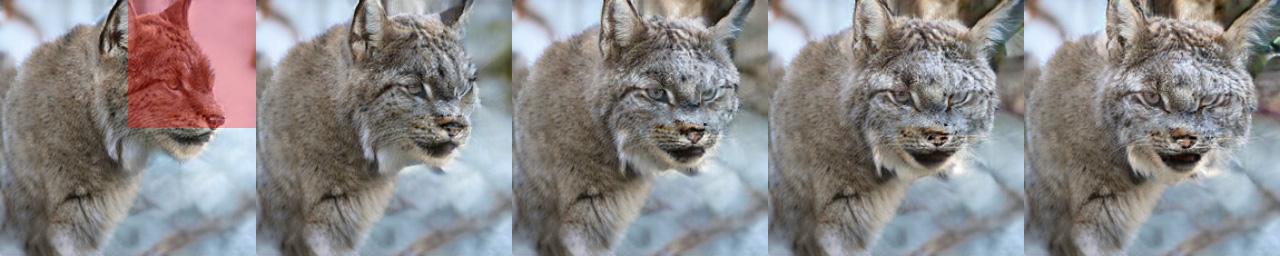}

\caption{Comparison between latent space interpolation (the top row in each panel) and feature blending (the bottom row in each panel), based on the same pair of latent variables. Feature blending was applied to the feature maps at the varying intensity in the red regions (higher intensity in the right).}
\label{fig:exp-z_vs_fb}

\end{figure}

Figure \ref{fig:exp-feature-collage} shows the examples of our feature blending.
We are succeeding in making semantic modifications like ``changing the posture of a dog", ``changing the style of a part of an anime character" without destroying the quality of the original image.
We also can see that the performance is robust against the alignment of the region to be transformed.
We emphasize that we are succeeding in creating these transformations without using any information about the attributes to be altered. 
Also note the difference in the results between our method and the simple interpolation in the latent space (figure \ref{fig:exp-z_vs_fb}).
As we can see in the result, our method can maintain the semantic context outside the specified region.
We see that we are succeeding in controlling the region to be modified with the user-specifiable map of mixing coefficients.

\subsection{The Effect of Collaging at Each Layer}
\label{exp:layers}

\begin{figure}[t]
\setlength{\abovecaptionskip}{0.10in}

\includegraphics[width=3.25in]{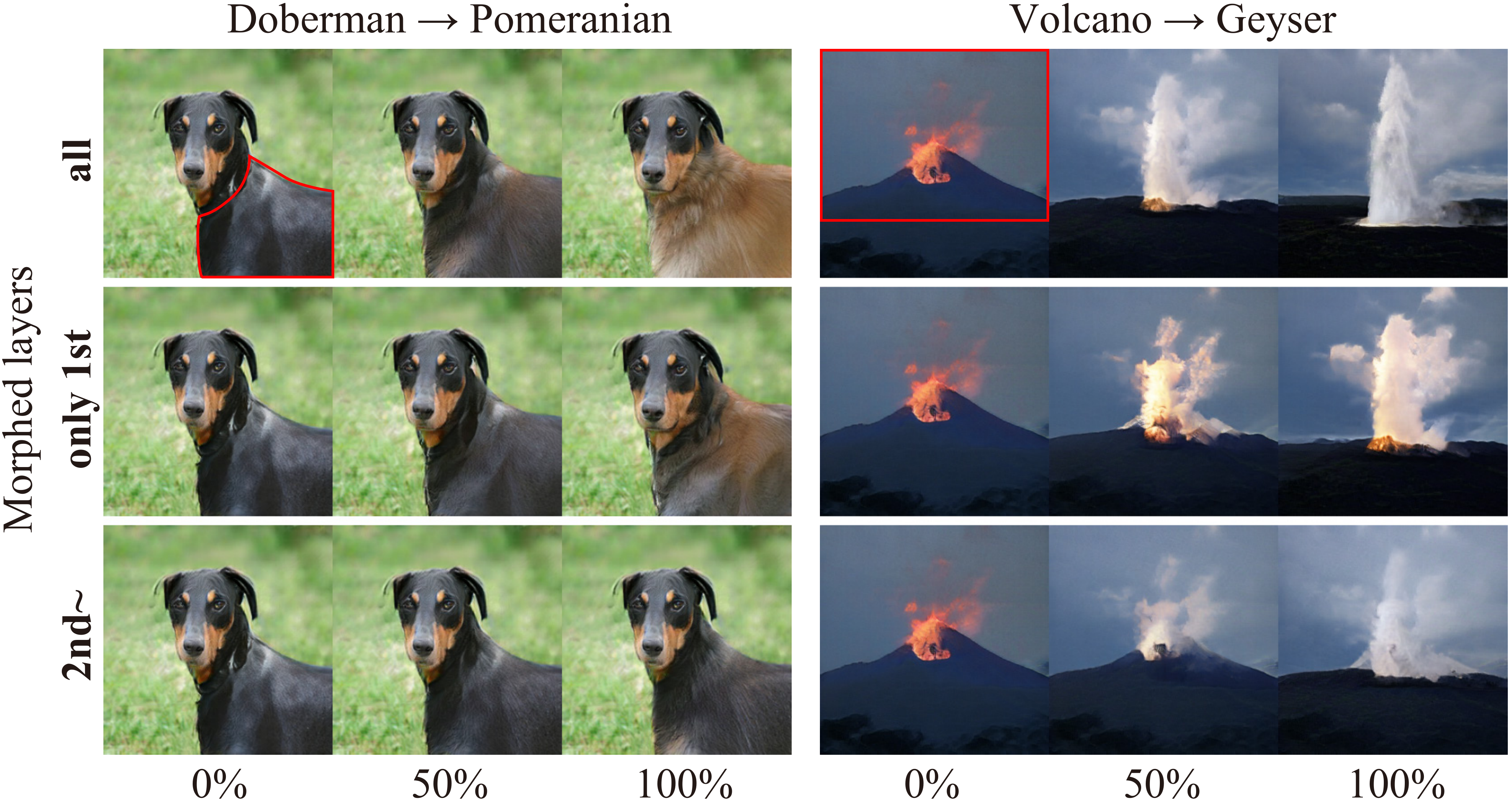}

\caption{Effect of collaging at different layers. From top to bottom, the results of applying the sCBN to all layers~(first row), just to the layer closest to the input~(second row), and to all layers except the first layer~(third row). Exclusive modification of the first layer tends to affect just the global feature of the image. }
\label{fig:layer-class}

\end{figure}

We conducted a set of ablation studies to investigate the influence of the modification at each layer.
The images in the figure~\ref{fig:layer-class} are results obtained by applying sCBN to (1) all layers, (2) the layer closest to the input (first layer), and (3) all the layers except the first layer.
As we mentioned in the previous section, the layers closer to $z$ tend to influence more global features, and the layers closer to $x$ tend to influence more local features.
As we can see in the figure, the features affected by the manipulation of the layers close to $z$ (body shape of dog, ridge line of mountain) are somewhat semantically independent from the features affected by the layers close to $x$ (fur texture, local features of ``lava").

\begin{figure}[t]

\centering
\includegraphics[width=3.2in]{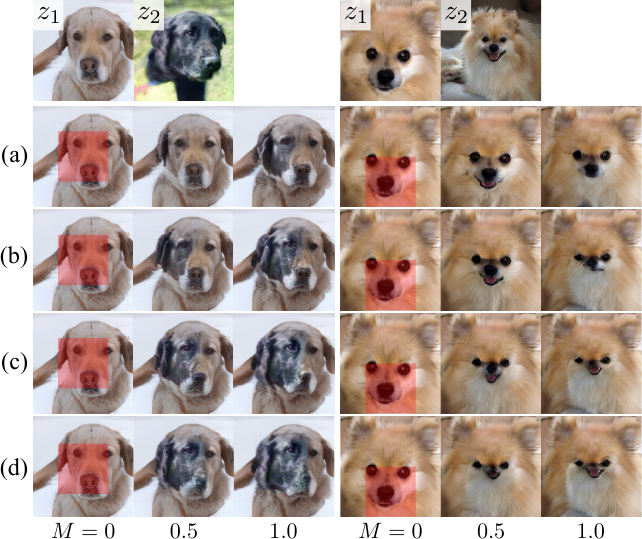}

\caption{Results of applying the feature blending to different sets of layers (a) $\ell$=1, (b) $\ell$=1,2, (c) $\ell$=1,2,3, and (d) $\ell$=1,2,3,4 with different mixing weight in the red-shaded regions. $M=\alpha$ means that the blending ratio of $z_1$ and $z_2$ in the red shaded region is $1-\alpha:\alpha$.}
\label{fig:layer-feature}

\end{figure}

Figure~\ref{fig:layer-feature} shows the result of applying the feature blending to different layers ($\ell \in 1,2,3,4$) with different blending weights.
When the blending is applied exclusively to the layer $\ell = 1$~(closest to the input), the global features like ``body shape of a dog" is affected, and local features like ``fur textures" are preserved..
We observe the opposite effect when the blending is applied to the layers closer to $x$. 
Also, when the reference image is significantly different from the target image in terms of its topology, the exclusive modification to the layers close to $x$ tends to produce artifacts.

These results suggest that, if necessary, we may customize our method to more finely control the locality and the intensity of the transformation by applying our methods with carefully chosen mixture-of-coefficient maps to each layer.

\subsection{Real Image Transformation}

In order to provide some quantitative measurement for the quality of our real image transformation, we evaluated the fidelity of the transformed images with classification accuracy and human perceptual test.
We applied our sCBN to the real images in ImageNet and conducted the translation of 
(1) cat $\to$ big cat, (2) cat $\to$ dog, and (3) dog $\to$ dog, and evaluated the classification accuracy of the transformed images by an inception-v3 classifier trained on ImageNet for the classification of 1000 classes.
For each pair of image-domains, we selected four classes from both source and target domains,
and conducted 1000 tasks of translating an image of a randomly selected source class to a randomly selected target class (\eg, one feline species to one canine species).
We used UNIT \cite{liu2017unsupervised} and MNUIT \cite{huang2018multimodal} as  baselines.
Because MUNIT is not designed for class-to-class translation, we conducted MUNIT using the set of images in the target class as the reference images.
Table \ref{tab:fidelity} summarizes the result.
For each set of translation task, our method achieved better top-5 error rate than the other methods.
This result confirms the efficacy of our method of real image transformation.
\setlength\textfloatsep{4mm}

\begin{table}[t]
\setlength{\abovecaptionskip}{0.10in}
\centering
\begin{tabular}{cccc}
\hline
Method & cat $\to$ big cat & cat $\to$ dog & dog $\to$ dog \\
\hline
Ours & 7.8\% & 21.1\% & 20.8\% \\
\hline
UNIT & 14.8\% & N/A & 36.2\% \\
\hline
MUNIT & 26.0\% & 55.4\% & N/A \\
\hline
\end{tabular}
\caption{Comparison of top-5 category classification error rate after class translation between two domains.}
 \label{tab:fidelity} 
\end{table}
We used Amazon Mechanical Turk (AMT) for a human-perceptual test. 
We asked each AMT worker to decide if the transformed images produced by our method are more photo-realistic than the images produced by MUNIT/UNIT.
For each comparison, we used 200 AMT workers for the evaluation and asked each worker to make votes for 10 pair of images. 
Table \ref{tab:amt} is the summary of the result.  
We see that our method outperforms both MUNIT and UNIT for the translations in all pairs of domains.
Our method is also capable of many-to-many translation over a set of 100 or more classes (see the supplementary material).

\begin{table}[t!]
\centering
\begin{tabular}{cccc}
\hline
vs. & cat $\to$ big cat & cat $\to$ dog & dog $\to$ dog \\
\hline
UNIT & $83.9$\% & N/A & $87.0$\% \\
\hline
MUNIT & 75.4\% & $89.7$\% & N/A \\
\hline
\end{tabular}
\vspace{0.5mm}
\caption{The rates at which human-agents considered the images produced by our algorithm to be more photorealistic than the rival methods (UNIT, MUNIT). Each rate is computed over 2000 pairs of a translation result and a human-individual. Chance is at 50\%.} 
 \label{tab:amt} 
\end{table}

\section{Conclusions}

By construction, our method is limited by the ability of the used generator.
Our method simply cannot make expressions beyond the representation power of the used generator.
This limitation is particularly true when we apply our method together with manifold projection to transform a real image.
For example, our method is likely to perform poorly in the transformation of the image of a specific individual unless the used generator is capable of accurately reconstructing the face of the target individual.  
The process of manifold projection \textit{projects} the target image to the restricted space of images that are reconstructable by the used generator, and some information is bound to be lost in the process.

We shall also make some remark about the weakness of our feature blending.
As we described throughout the text, the main strength of our feature blending is that it does not require the user to provide the explicit definition of the feature to be blended. 
It is very natural that this advantage comes at the cost of blending the feature that is not intended to be blended.  
Because we ask the user to specify just the \textit{source region} from which to extract the feature,  \textit{all features} that are contained in the user-specified region become subject to the blending process. 
We may be able to further fine-tune the image synthesis by using the method in \cite{bau2018gan} to identify the specific unit in the internal feature space that corresponds to the target feature to be transferred.   

Also, conditional normalization layers are capable of handling not only class conditions but other types of information like verbal statements as well.
One might be able to conduct an even wider variety of spatial transformation by making use of them. 
Applications of our method to non-image dataset is also an interesting direction for future work. 

\paragraph{Acknowledgements}

We would like to acknowledge Jason Naradowsky, Masaki Saito, Jin Yanghua, and Atsuhiro Noguchi for helpful insights and suggestions.

{\small
\bibliographystyle{ieee}
\bibliography{egbib}
}

\clearpage
\appendix

\section{Implementation in Various GANs}

Our label collaging method can be applied to any GAN equipped with conditional normalization layers (\eg, CBN, AdaIN) by replacing the layers with their spatial variants.
Our feature collaging method can be applied to any GAN as well by specifying the intermediate feature maps $\{F^{(\ell)}\}$ to be manipulated.
For the ResNet-based architectures like SNGAN and BigGAN, we chose to manipulate the output feature map and the first 4$\times$4 feature map of each ResBlock.
For StyleGAN, we chose to manipulate the output of each Synthesis Block. 
Each Synthesis Block consists of two AdaIn layers, one upsampling operator and one convolution layer. 

\section{Fast Manifold Projection}

In order to enable the semantic modification of real images in \textit{real time}, we explored a way to speed up the manifold projection step. 
Our implementation was able to speed up the process by more than 30\%.

\subsection{Latent Space Expansion with Auxiliary Network}
\begin{figure}[t]
\centering
\includegraphics[width=3.25in]{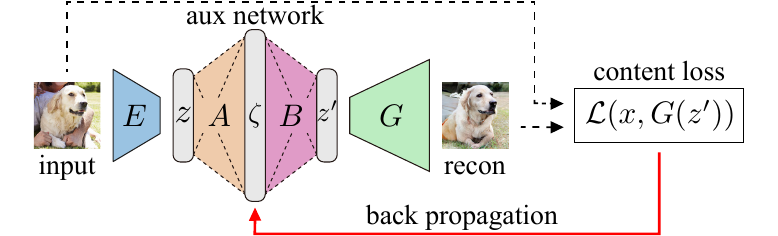}
\caption{An overview of our manifold projection algorithm employing an auxiliary network ($A + B$).}
\label{fig:aux}
\end{figure}

Image features are generally entangled in the latent space of a generative model, and the optimization on the complex landscape of loss can be time-consuming.
To speed up the process, we construct an auxiliary network that embeds $z$ in higher dimensional space.
The auxiliary network consists of an embedding map $A$ that converts $z$ into high dimensional $\zeta$, and a projection map $B$ that converts $\zeta$ back to $z$ (figure \ref{fig:aux}).
That is, instead of calibrating the latent variable $z$ by backpropagating $\mathcal{L}$ through $G$, we will calibrate $\zeta$ by backpropagating $\mathcal{L}$ through $G$ and $B$.
The goal of training this auxiliary network is to find the map $B$ that allows us to represent the landscape of $\mathcal{L}$ in a form that is more suitable for optimization, together with the map $A$ that can embed $z$ in a way that is well-suited for the learning on the landscape.
Let $\zeta_j$ be the variable in the high dimensional latent space after $j$ rounds of calibration.
The update rule of $\zeta_j$ we use here is
\begin{align}
    &\zeta_{j+1} = \zeta_j - \alpha_{j} \nabla_{\zeta_j} \mathcal{L}(G(B(\zeta_j), x),~\zeta_0=A(E(x)),
\end{align}
where $\alpha_j$ is the length parameter at the $j$-th round.
We train the networks $A$ and $B$ using the following loss function:
\begin{align}
  \mathcal{L}_{\mathrm{AB}} &:= \sum_{j=0}^{N_{\mathrm{iter}}} w_j \mathcal{L}(G(B(\zeta_j)), x) +\lambda \| B(A(z)) - z\|_2^2
\end{align}
where constants $\{w_j\}$ determine the importance of the $j$-th round of the calibration. 
For the first term, $A$ and $B$ are updated through the backpropagation from $\zeta_j$.
The second term makes sure that $z$ can be reconstructed from $\zeta$.
This process can be seen as a variant of meta-learning \cite{finn2017model,munkhdalai2018rapid}.

For the auxiliary maps $A$ and $B$, we used a $5$-layer MLP with hidden units of size $(1000, 10000, 1000)$ equipped with a trainable PReLU activation function at respective layers, and treated the 10000-dimensional layer as the extended latent variable $\zeta$.
For the evaluation of $\mathcal{L}_{\mathrm{AB}}$, we used $N_{\mathrm{iter}}=2$, $w_0=20.0, w_1=2.0, w_2=1.0$, and $\lambda=100.0$.

\subsection{Training Procedure}
\begin{figure}[t]
\centering
\includegraphics[height=2.0in]{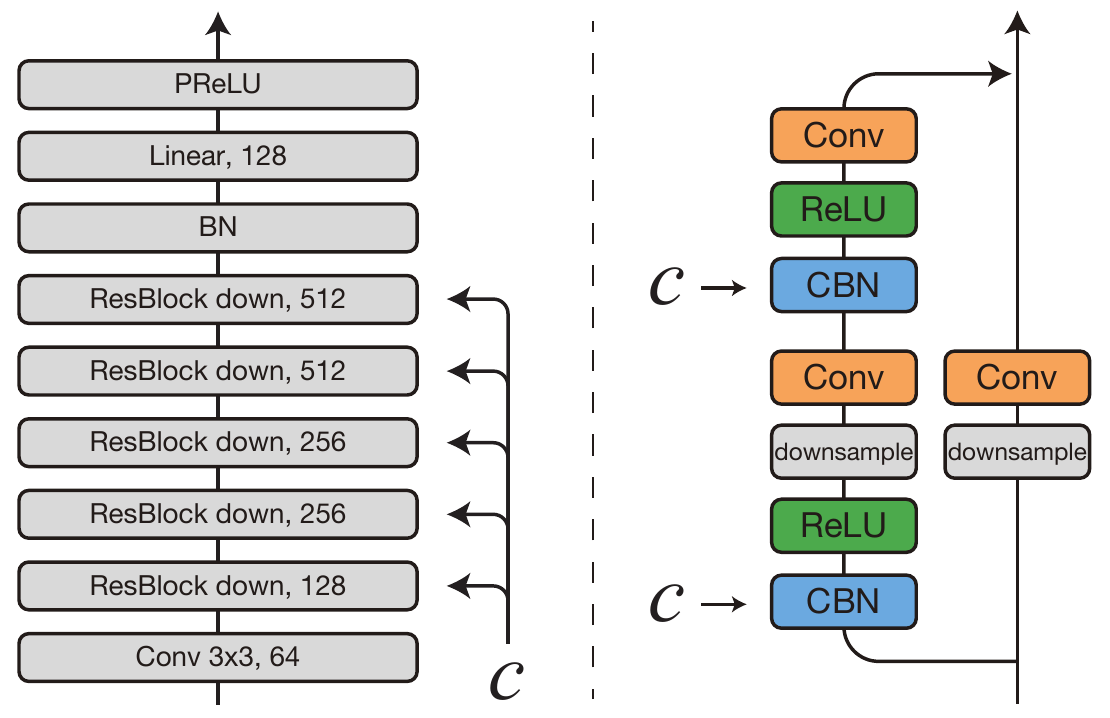}
\caption{Left: The architecture of the encoder for manifold projection.  The model is identical to the discriminator of SNGAN except that its output is a 128-dimensional latent vector $z=E(x)$ instead of a scaler value. Right: The structure of the ResBlock used in the encoder. } 
\label{fig:encoder}
\end{figure}

We trained the three components of the model ($G+D$, $E$, $A+B$) separately, in order. 
We first trained a pair of generator and discriminator by following the procedure of conditional GANs described in \cite{miyato2018spectral,miyato2018cgans}. 
We then trained the encoder network (figure~\ref{fig:encoder}) for the trained generator using the objective function defined based on the trained discriminator (see the main section in the article about the manifold projection).  Finally, fixing the encoder and the generator we trained, we trained the auxiliary networks to enhance the manifold-embedding optimization.

For the training of the auxiliary network, we used AdaGrad with adaptive learning rate for the gradient descent to calculate $\{\zeta_i\}_{i=1}^{N_\mathrm{iter}}$.
We used AdaGrad for this procedure because other methods (\eg, Adam) causes numerical instability in double backpropagation for network parameter update.
We trained each model over $450,000$ iterations. For the training of auxiliary networks, we applied the gradient clipping with the threshold of $100.0$ and the weight decay at the rate of $0.0001$.
We trained the networks on a GeForce GTX TITAN X for about a week for each component.

\subsection{Speed-up by Latent Space Expansion}
\begin{figure}[t]
\setlength{\abovecaptionskip}{0.10in}
\centering
\includegraphics[width=3.25in]{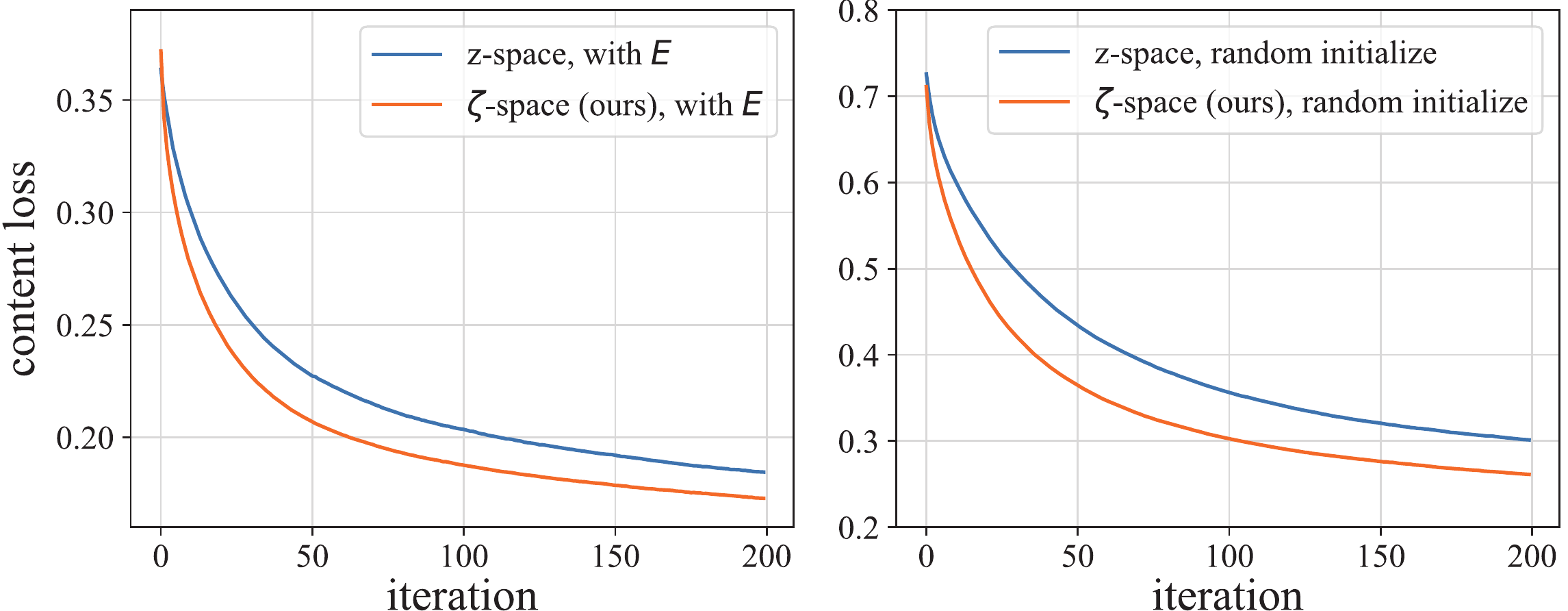}

\caption{Loss transition during optimization on $z$-space and $\zeta$-space using the encoder (left) and random initialization (right).}
\label{fig:optimization}

\end{figure}

We conducted an experiment to verify that our auxiliary network can indeed speed up the optimization process.  We used the DCGANs trained on the dog+cat dataset and evaluated the average loss for 1000 images randomly selected from ImageNet.  The optimization was done with Adam, implemented with the best learning rate $\alpha$ in the search-grid that achieved the fastest loss decrease. 
Figure~\ref{fig:optimization} compares the transition of the loss function learned on $z$-latent space against the loss transition produced on $\zeta$-latent space.  
To decrease the loss by the same amount, the learning of $\zeta$ required only less than $2/3$ the number of iterations required by the learning of $z$.
Calculation overhead due to the latent space expansion was negligible ($<1$\%).
Indeed, the learning process depends on the initial value of the latent variable $z$ or $\zeta$.  
When compared to random initialization, the optimization process on both $z$ space and $\zeta$ space proceeded faster when we set the initial value at $E(x)$ and $E(A(x))$.
This result is indicative of the importance of training $A$.
We also studied the robustness of our manifold projection algorithm against the choice of the hyper-parameter.
We performed optimization with various learning rates $\alpha$ of Adam for updating $\zeta$.
Our study suggests that our algorithm is quite robust with respect to the learning rate.

\begin{figure}[t]
\centering
\includegraphics[width=2.4in]{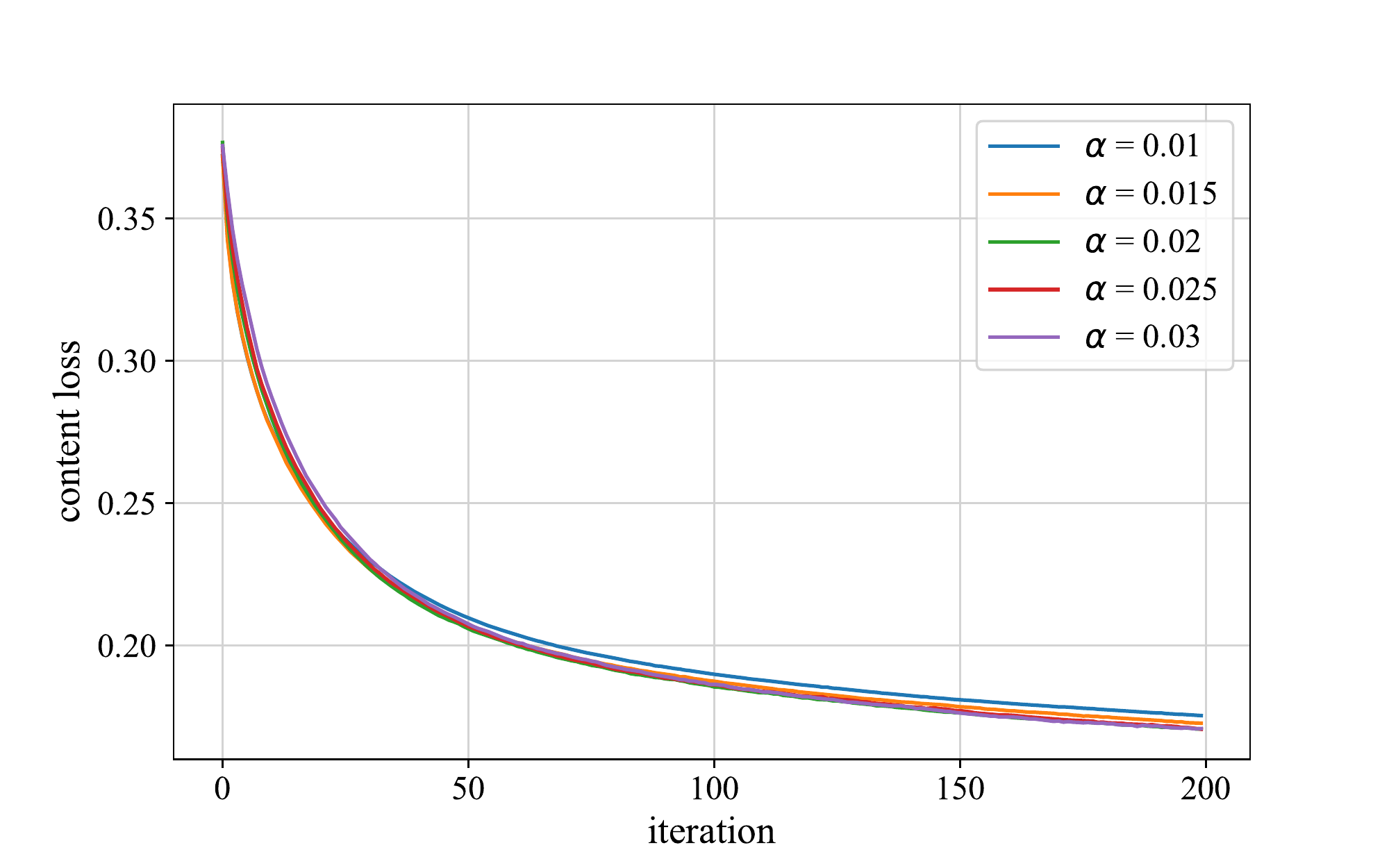}

\caption{Loss transition of manifold projection by optimization on $\zeta$-space with various learning rates $\alpha$ of Adam.}
\label{fig:optimization_alpha}

\end{figure}

\section{Internal Representation Collage vs. Pixel-space Collage}
\begin{figure}[t]
\centering
\includegraphics[width=3.2in]{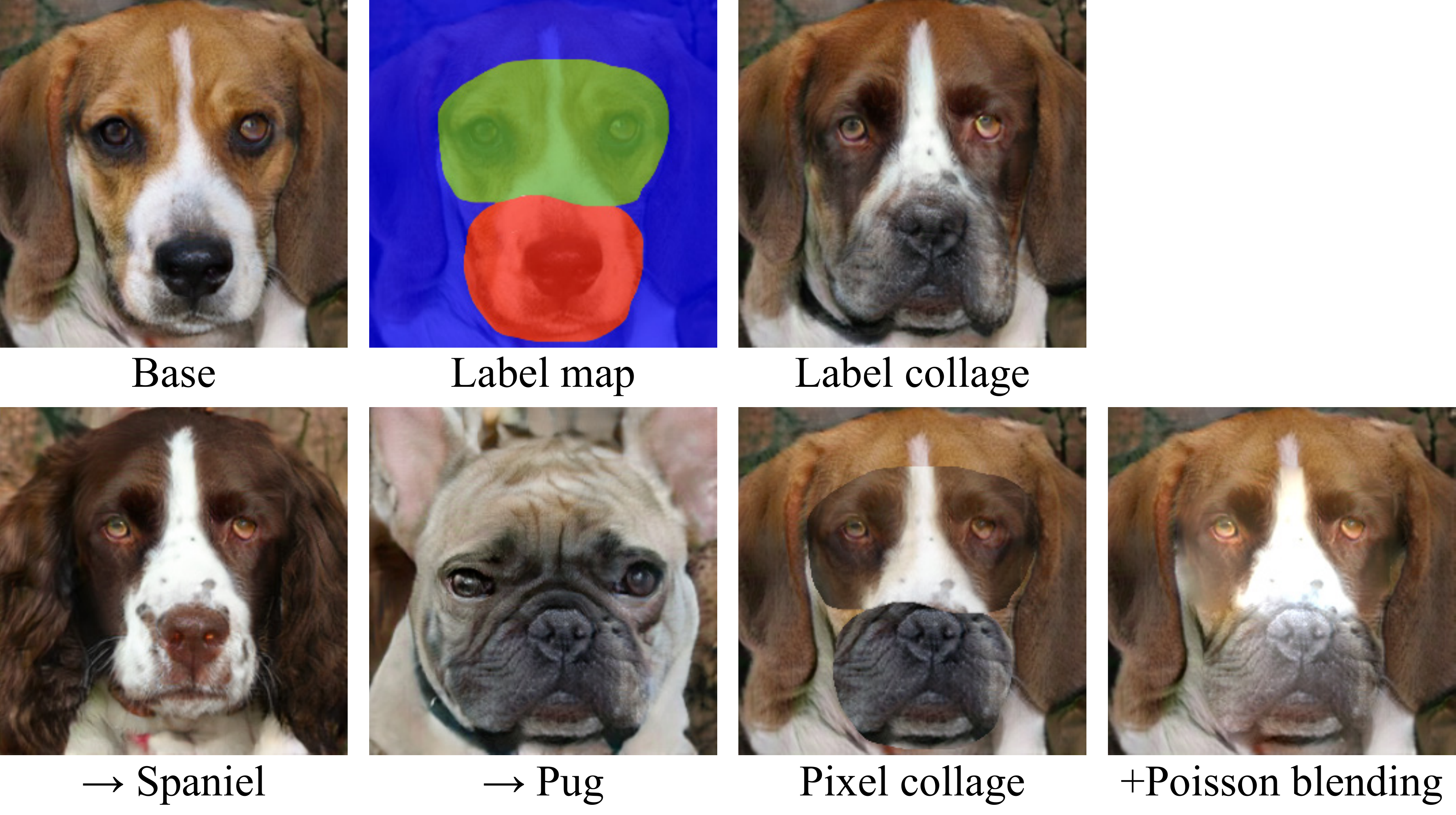}\\
\vspace{0.1in}
\includegraphics[width=3.2in]{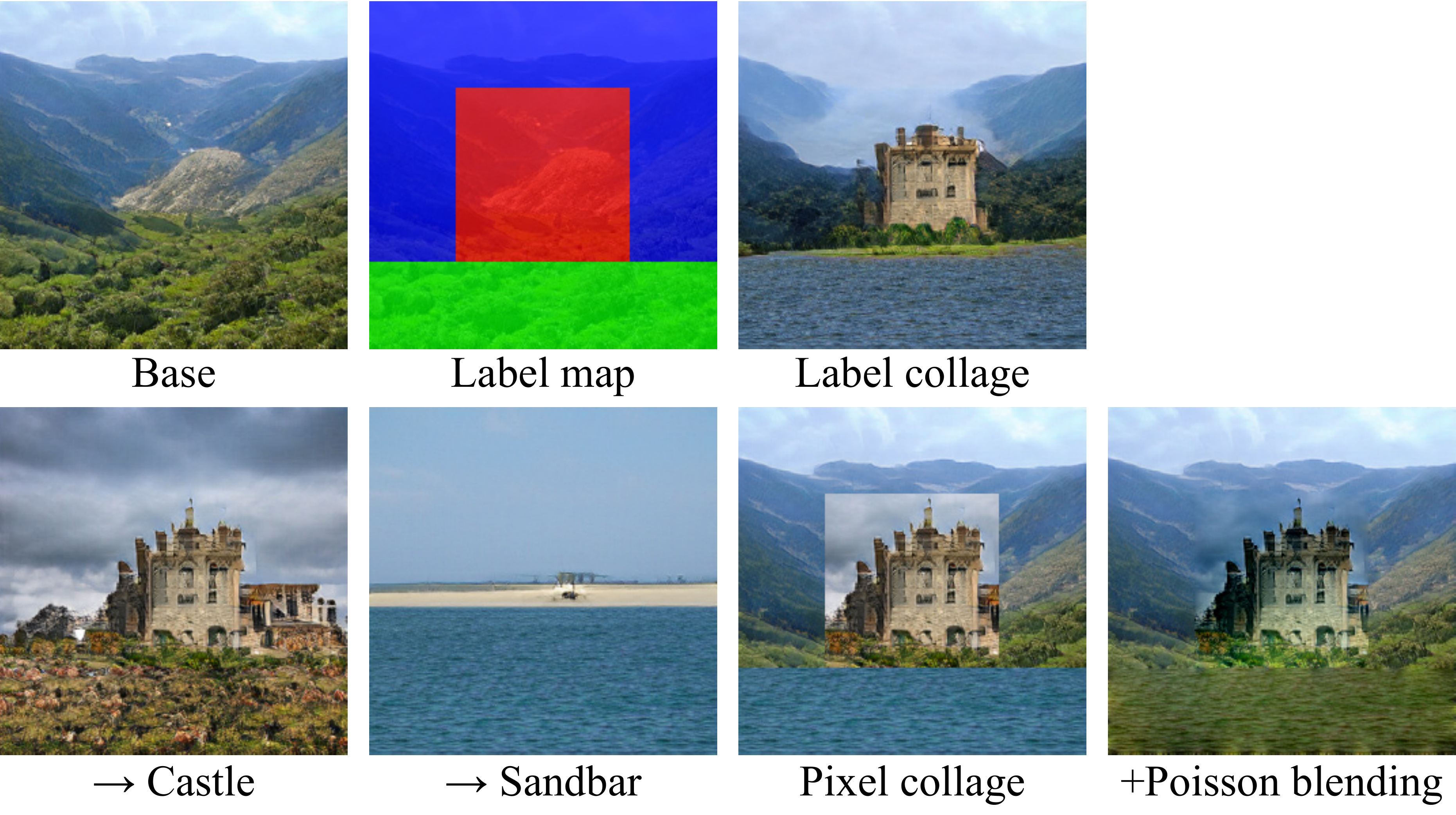}

\caption{Pixel space collaging (naive copy-and-paste) vs Latent space label collaging on generated images (top panel:SNGAN, bottom panel:BigGAN). The label information in the green colored region and the red colored region of the base image are changed. Our method automatically adjusts the spatially varying intensity of the modification and render an image that is natural over all regions.  Pixel space collaging is unnatural even after the application of Poisson blending, especially at the boundary of the modification.}

\label{fig:label_vs_pixel}

\end{figure}

Our methods modify the target image by manipulating its intermediate representation in the generator's feature space. 
The most prominent advantage of this strategy is that it allows the generator to automatically adjust the spatially varying intensity of the modification to make the final output \textit{natural}.
In this section, we compare the results of our method against the most basic collaging method of modifying the image in the pixel space \textit{pixel collage}. 
The images in the Figure~\ref{fig:label_vs_pixel} are the results of label collaging for an image obtained from the same latent variable. 
As we see in Figure~\ref{fig:label_vs_pixel}, the result of pixel collage is deformed/blurry even after the post-processing of Poisson blending, and the boundary of modified regions is unnatural.
On the other hand, our method is automatically adjusting the intensity to naturally match the modified region and the neighboring region  (\eg, mouth shape, shoreline).

\begin{figure}[t]
\centering
\includegraphics[width=3.25in]{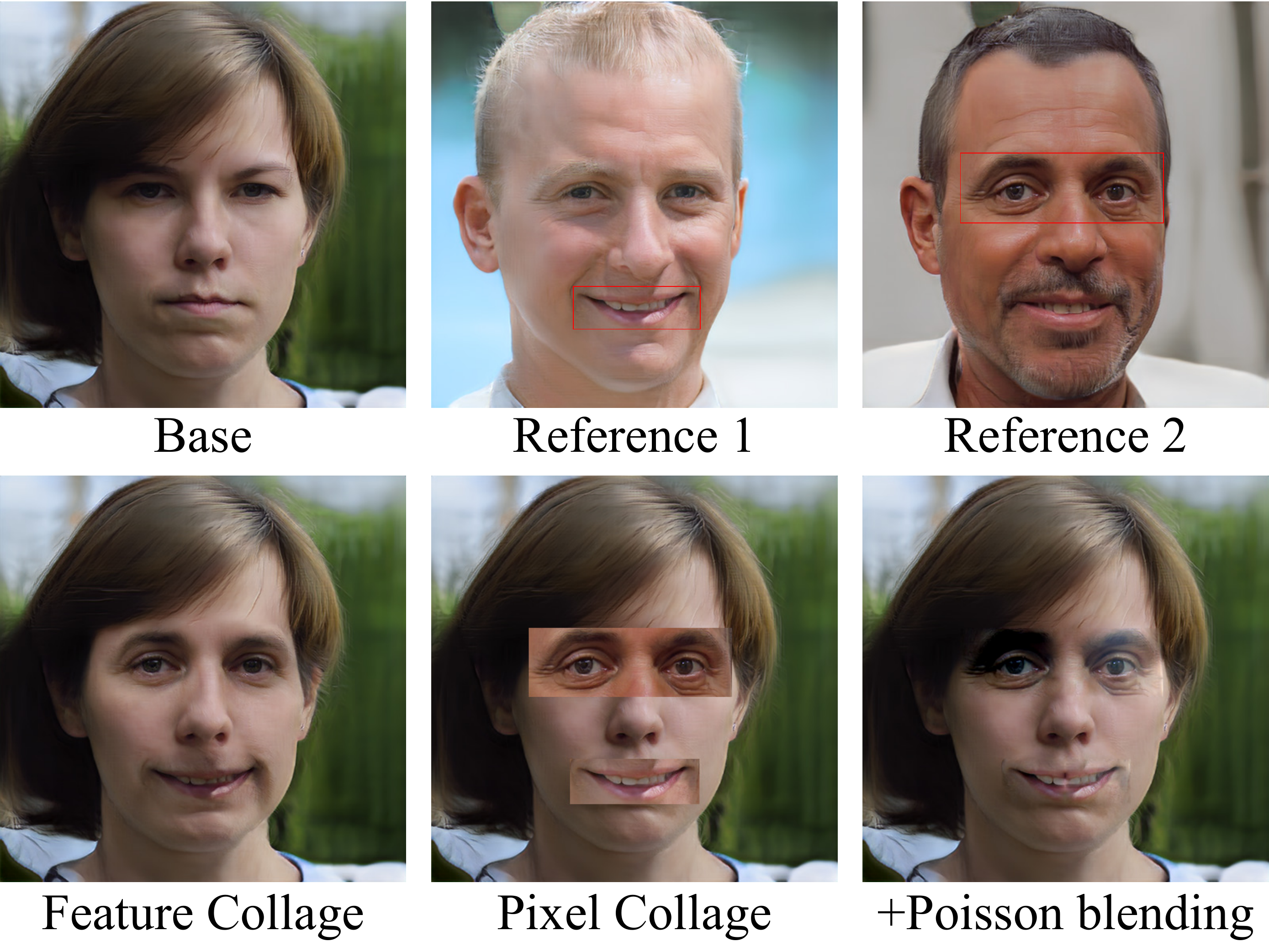}\\
\vspace{0.1in}
\includegraphics[width=3.25in]{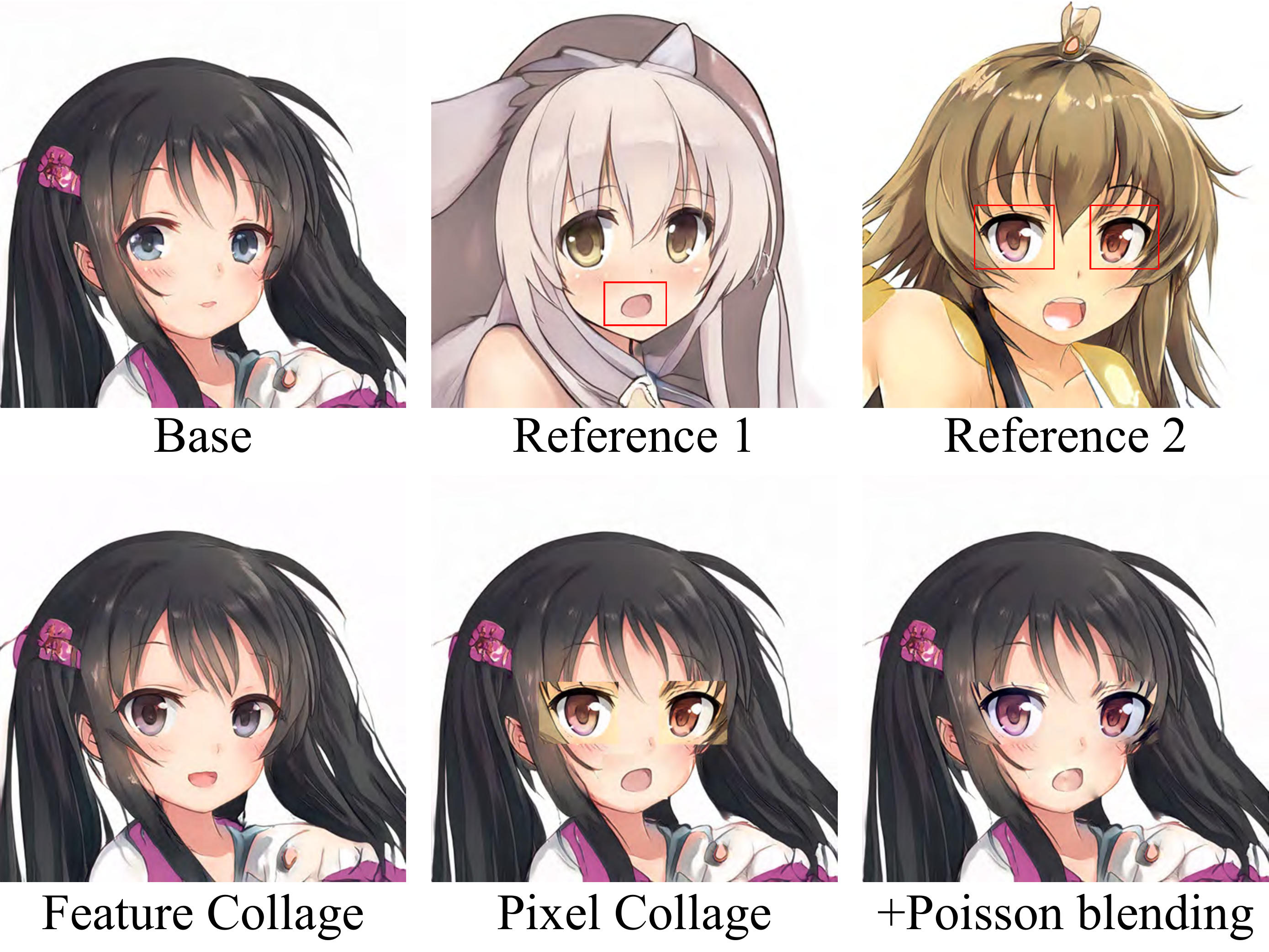}

\caption{Pixel space collaging vs latent space feature collaging on StyleGAN generated images.  Red-framed regions in the reference images are blended (transplanted) into the base image.  Notice that our method is rendering much more natural images than the naive collaging, especially at regions surrounding the modified areas. } 
\label{fig:feature_vs_pixel}

\end{figure}

Figure~\ref{fig:feature_vs_pixel} compares our feature collaging against the naive pixel space collaging. 
Notice that our method is not creating any artifacts around the modified regions. 
On the other hand, the image created by pixel space collaging is strongly deformed/blurry around the region of modification, even after the application of Poisson blending.

\section{Real Image Transformation Study}

\begin{figure}[t]
\centering
\includegraphics[width=3.1in]{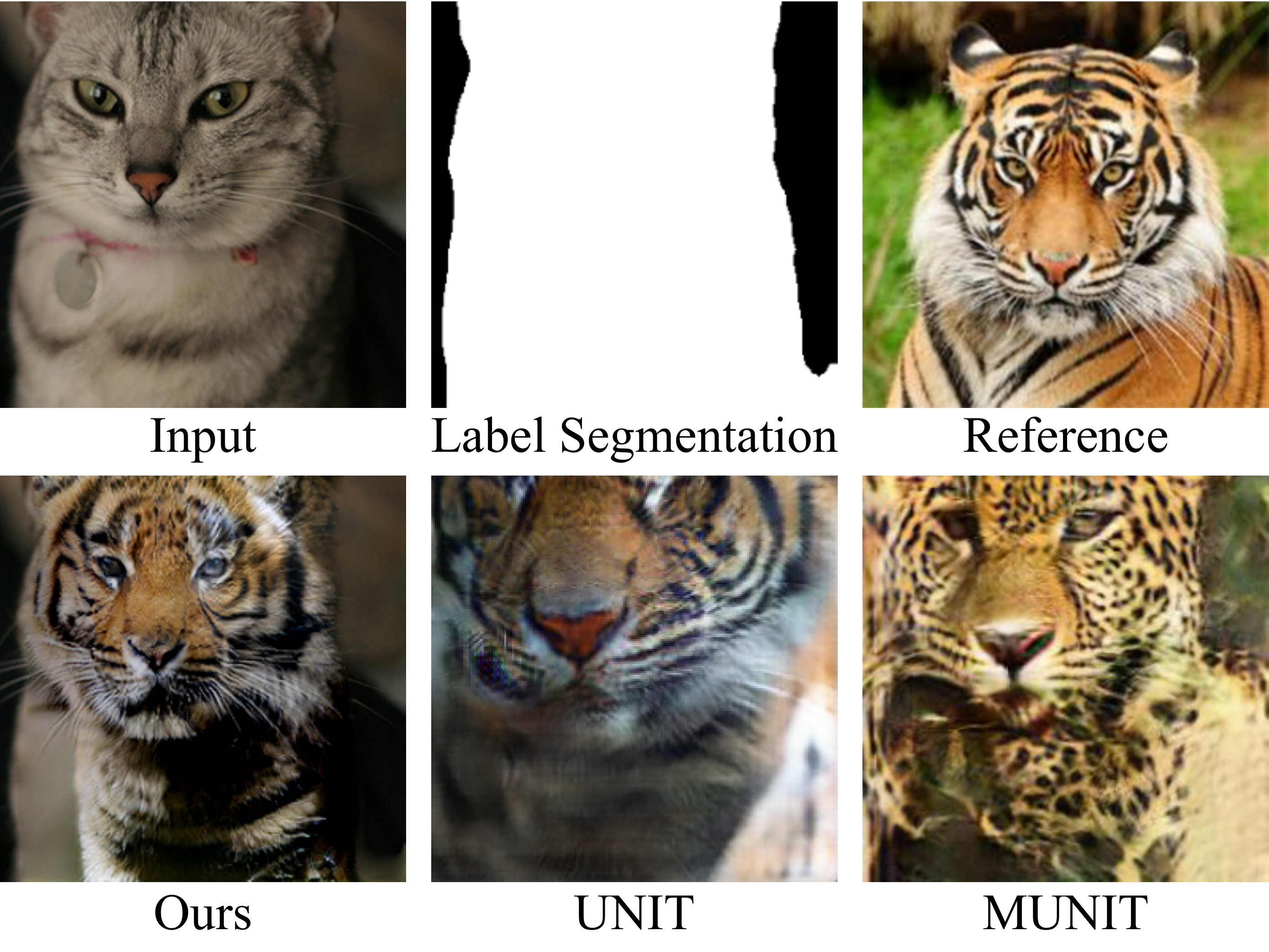}

\caption{Comparison of our methods against MUNIT and UNIT for the transformation of real images. In the ablation study, the reference image was used by MUNIT to specify the target class.}
\label{fig:fidelity-example}

\end{figure}

For evaluating the fidelity of real image transformation, we conducted a set of automatic spatial class translations.
For each one the selected images, we  
(1) used a pre-trained model to extract the region of the object to be transformed (dog/cat), (2) conducted the manifold projection to obtain the $z$, 
(3) passed $z$ to the generator with the class map corresponding to the segmented region, and 
(4) conducted post-processing over the segmented region. 
For the semantic segmentation, we used a TensorFlow implementation of DeepLab v3 Xception model trained on MS COCO dataset\footnote{\url{https://github.com/tensorflow/models/tree/master/research/deeplab}}.
See Figure \ref{fig:fidelity-example} for the transformation examples.

\section{Global Translation of Real Image}

\begin{figure*}
\begin{center}
\includegraphics[width=17.4cm]{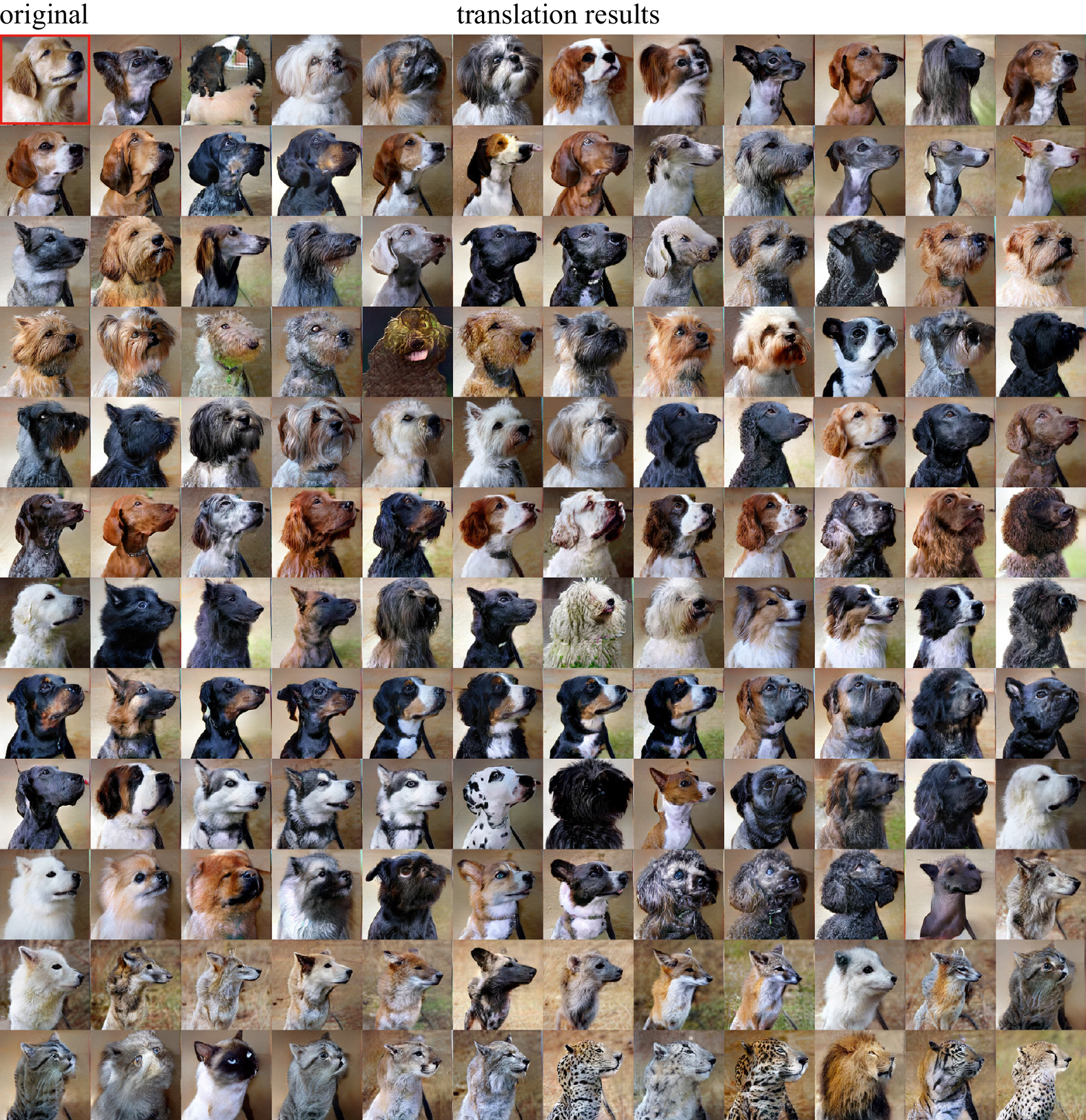}
\end{center}
\caption{One-to-many \textit{non-spatial} class-translation result. The most upper left image is the input sampled from the validation dataset of ImageNet, and the rest images are translation results to all the 143 dog+cat classes of ImageNet.
All the translation results are produced using a same latent variable calculated by the proposed algorithm.}
\label{fig:non-spatial-to-all}
\end{figure*}

In order to verify the sheer ability of our translation method,  we conducted a translation task for the entire image as well (as opposed to the translation for a user-specified subregion). For each of the selected images from the ImageNet, we calculated a latent encoding variable $z$, and applied a set of spatial uniform class-condition $c$ to the layers of the decoder.
Figure~\ref{fig:non-spatial-to-all} contains the translation of a real image (designated as \textit{original}) to all 143 object classes that were used for the training of the dog+cat model. We can see that the semantic information of the original image is naturally preserved in most of the translations.
\end{document}